\documentclass[11pt]{article}
\usepackage{algorithm}
\usepackage{algorithmicx}
\usepackage{algpseudocode}
\usepackage{amsmath}
\usepackage{booktabs}
\usepackage{multirow}
\usepackage{paralist}
\usepackage[final]{acl}
\usepackage{placeins}  
\usepackage{array}  
\usepackage{caption}
\usepackage{times}
\usepackage{latexsym}

\usepackage[most]{tcolorbox}      
\tcbuselibrary{breakable}      
\usepackage{multicol}          
\usepackage{graphicx}          
\usepackage{caption}           
\usepackage{amsmath, amssymb}  
\usepackage{hyperref}
\usepackage[utf8]{inputenc} 
\usepackage[T1]{fontenc}
\usepackage{subcaption}
\usepackage[utf8]{inputenc}
\usepackage{microtype}

\usepackage{inconsolata}

\usepackage{amsmath}
\usepackage{amssymb}
\usepackage{graphicx}
\usepackage{colortbl}  
\usepackage{xcolor}     
\usepackage{adjustbox}
\usepackage{cuted}
\usepackage{mathtools}
\usepackage{listings}

\title{FinHEAR: Human Expertise and Adaptive Risk-Aware Temporal Reasoning for Financial Decision-Making}

\author{
Jiaxiang Chen\textsuperscript{1}\thanks{Equal contribution.} \quad
Mingxi Zou\textsuperscript{1,2}\footnotemark[1] \quad
Zhuo Wang\textsuperscript{1} \quad
Qifan Wang\textsuperscript{4} \\
\textbf{
Dongning Sun\textsuperscript{3} \quad
Chi Zhang\textsuperscript{2}\thanks{Corresponding author.} \quad
Zenglin Xu\textsuperscript{1,5}\footnotemark[2]
} \\
\textsuperscript{1}Fudan University \quad
\textsuperscript{2}Tensorpacific \quad
\textsuperscript{3}Peng Cheng Laboratory \quad
\textsuperscript{4}Meta AI \\
\textsuperscript{5}Shanghai Academy of AI for Science (SAIS)\\
\texttt{\{jiaxiangchen23, 24210240447\}@m.fudan.edu.cn}
}

\begin{document}
\maketitle

\begin{abstract}

Financial decision-making presents unique challenges for language models, demanding temporal reasoning, adaptive risk assessment, and responsiveness to dynamic events. While large language models (LLMs) show strong general reasoning capabilities, they often fail to capture behavioral patterns central to human financial decisions—such as expert reliance under information asymmetry, loss-averse sensitivity, and feedback-driven temporal adjustment.
We propose \textbf{FinHEAR}, a multi-agent framework for \textbf{H}uman \textbf{E}xpertise and \textbf{A}daptive \textbf{R}isk-aware reasoning. FinHEAR orchestrates specialized LLM-based agents to analyze historical trends, interpret current events, and retrieve expert-informed precedents within an event-centric pipeline. Grounded in behavioral economics, it incorporates expert-guided retrieval, confidence-adjusted position sizing, and outcome-based refinement to enhance interpretability and robustness.
Empirical results on curated financial datasets show that FinHEAR consistently outperforms strong baselines across trend prediction and trading tasks, achieving higher accuracy and better risk-adjusted returns.
The code\footnote{\url{https://github.com/pilgrim00/FinHEAR}} is now publicly accessible.
\end{abstract}


\section{Introduction}
\begin{figure}[tb] %
    \centering 
    \includegraphics[width=0.5\textwidth]{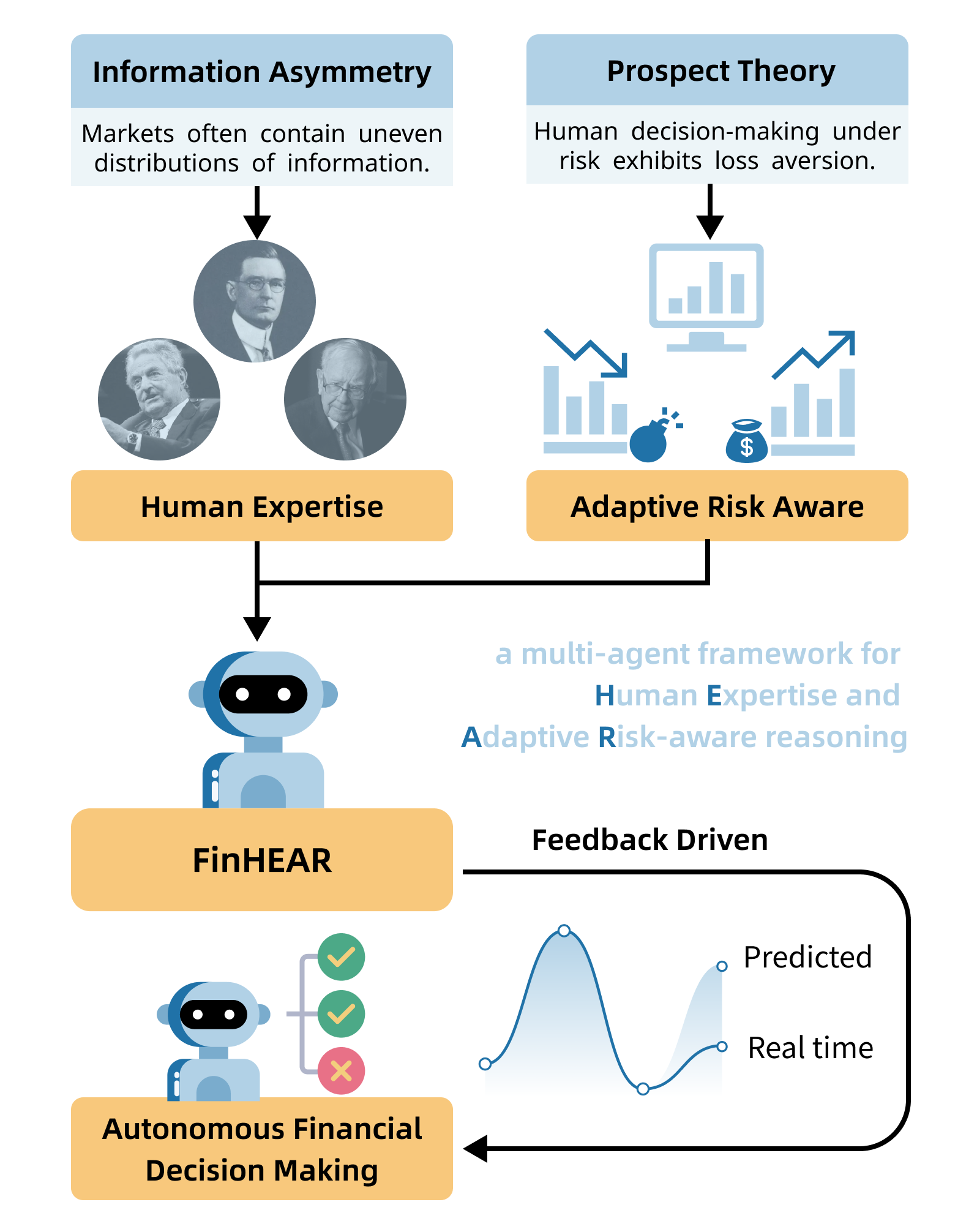} 
    \caption{Conceptual motivation behind FinHEAR. The framework is grounded in two behavioral finance principles—\textit{information asymmetry} and \textit{prospect theory}—which emphasize the role of expert knowledge and loss-averse risk behavior in financial decision-making.}
    \label{fig:illu}
\end{figure}

Large language models (LLMs)~\cite{gpt4,llama3.1} have demonstrated strong capabilities in general reasoning tasks, including chain-of-thought prompting~\cite{cot}, in-context learning~\cite{icl_survey}, and code understanding~\cite{code_understand}. However, applying LLMs to financial domains remains challenging due to the dynamic, uncertain, and event-driven nature of real-world markets. Effective financial decision-making requires temporal reasoning, adaptive risk assessment, and responsiveness to external shocks—far beyond the scope of static reasoning benchmarks.

These challenges are further amplified by the complexity of financial data, which is high-frequency, multi-source, and nonlinear~\cite{cao2022ai,Bahoo2024}. Crucially, financial decisions are not purely rational computations—they reflect behavioral patterns shaped by human cognition. Under uncertainty, individuals rely on expert heuristics~\cite{BoundedRationality,bounded,InformationAsymmetry1}, adapting decisions through bounded rationality and accumulated experience. Moreover, risk-sensitive behaviors such as loss aversion and feedback-based adjustment, as formalized in Prospect Theory~\cite{ProspectTheory,ProspectTheoryloss}, remain largely unmodeled in existing AI systems. Bridging this gap is essential for aligning LLM reasoning with real-world, high-stakes decision-making.


Prior work has explored rule-based systems, deep learning models, and more recently, LLM-based frameworks for financial decision-making. Rule-based approaches encode expert heuristics but struggle with generalization in dynamic, event-driven environments. Deep learning models, including LSTM~\cite{lstm}, Transformer~\cite{attention}, and their variants~\cite{alstm,informer}, improve temporal modeling but often lack interpretability and adaptivity. Reinforcement learning (e.g., PPO~\cite{ppo}) offers policy learning under uncertainty, yet remains data-hungry and hard to align with human intuition.

Despite these advancements, existing methods still fall short of key requirements for reliable financial decision-making. First, they lack structured mechanisms to integrate accumulated human expertise—critical for mitigating information asymmetry and supporting decisions under bounded rationality. Second, they exhibit limited sensitivity to risk, often ignoring the loss-averse and risk-adjusted behaviors emphasized by behavioral finance theories such as Prospect Theory. Third, they fail to ensure temporal consistency: unlike static reasoning tasks, financial decisions unfold over time, where early-stage errors can compound if not continually revised. Addressing these gaps is vital for building systems that are not only accurate, but also robust, interpretable, and adaptive to real-world market dynamics.


We introduce \textbf{FinHEAR}—a multi-agent framework for \textbf{H}uman \textbf{E}xpertise and \textbf{A}daptive \textbf{R}isk-aware temporal reasoning. FinHEAR addresses three key challenges in financial decision-making: incomplete information, risk sensitivity, and temporal consistency.To reduce information asymmetry, one agent retrieves relevant expert cases to emulate human-guided decisions under bounded rationality—mirroring how investors rely on heuristics under cognitive constraints. To enhance risk awareness, another agent estimates market risk via inter-agent disagreement and adjusts position sizes based on prediction confidence, modeling the loss-averse behavior described by Prospect Theory. A feedback mechanism further refines earlier predictions using realized outcomes, promoting temporal coherence and limiting error propagation.These components are embedded in a structured, event-driven pipeline, enabling FinHEAR to produce reasoning that is both effective and consistent with behavioral economic theories of human decision-making under uncertainty.

We construct an event-driven financial dataset linking multi-stock price movements with macroeconomic events, firm-level news, and expert commentary. Empirical results show that FinHEAR achieves substantial gains in both trend forecasting and decision making, consistently outperforming strong baselines across key metrics such as ACC, MCC, CR, SR, MDD and CalmarR.

Our key contributions are summarized as follows: 
\begin{compactitem}
\item We propose \textbf{FinHEAR}, a multi-agent reasoning framework that integrates \textbf{H}uman \textbf{E}xpertise and \textbf{A}daptive \textbf{R}isk-aware temporal reasoning within a structured, event-driven pipeline—the first to explicitly combine multi-agent coordination with behavioral decision principles for financial reasoning.

\item We design explicit mechanisms for expert-guided retrieval, risk-aware action modulation, and feedback-driven temporal refinement——grounded in behavioral decision theories including bounded rationality and loss aversion.

\item We validate FinHEAR on real-world financial tasks in both trend forecasting and decision making task, consistently outperforming strong baselines across many metrics.
\end{compactitem}

\section{Related Work}

\paragraph{Deep Learning Methods.}
Deep learning has been extensively applied to financial forecasting, with models such as LSTM~\cite{lstm}, GRU~\cite{gru}, and ALSTM~\cite{alstm} capturing sequential dependencies and emphasizing salient time steps. Transformer-based models~\cite{attention} improved long-range modeling, while Informer~\cite{informer} and TimesNet~\cite{wu2023timesnet} enhanced efficiency and temporal variation modeling. Extensions like StockNet~\cite{stocknet} integrated sentiment features, and reinforcement learning methods~\cite{ppo,rl_survey} introduced reward-driven decision-making, further refined by Logic-Q~\cite{li2025logic} with structured priors.

However, these models primarily focus on pattern extraction and lack mechanisms for event understanding and risk-adaptive reasoning—gaps that \textbf{FinHEAR} bridges through expert-informed retrieval, dynamic risk modeling, and feedback-based refinement.

\paragraph{LLM-Based Methods.}
Recent advances in large language models (LLMs) have enabled structured financial reasoning through prompting and memory-augmented techniques. Methods like Chain-of-Thought~\cite{cot,sc_cot,auto-cot}, ReAct~\cite{react}, and their extensions~\cite{tot,beats,fot,hiar-icl,booststep} promote stepwise reasoning and agent coordination. FinMEM~\cite{finmem} and FinCon~\cite{fincon} incorporate memory and multi-agent structures for risk-aware decision-making, while TradingAgents~\cite{tradingagents} mimics real-world organizational hierarchies to improve trading strategies.

While these methods introduce reasoning and coordination capabilities, \textbf{FinHEAR} advances further by unifying expert-guided retrieval, dynamic risk modeling, and feedback-based refinement in a behavior-aware framework.

\section{Our Method}
\begin{figure*}[htb]
    \centering
    \includegraphics[width=1.0\textwidth]{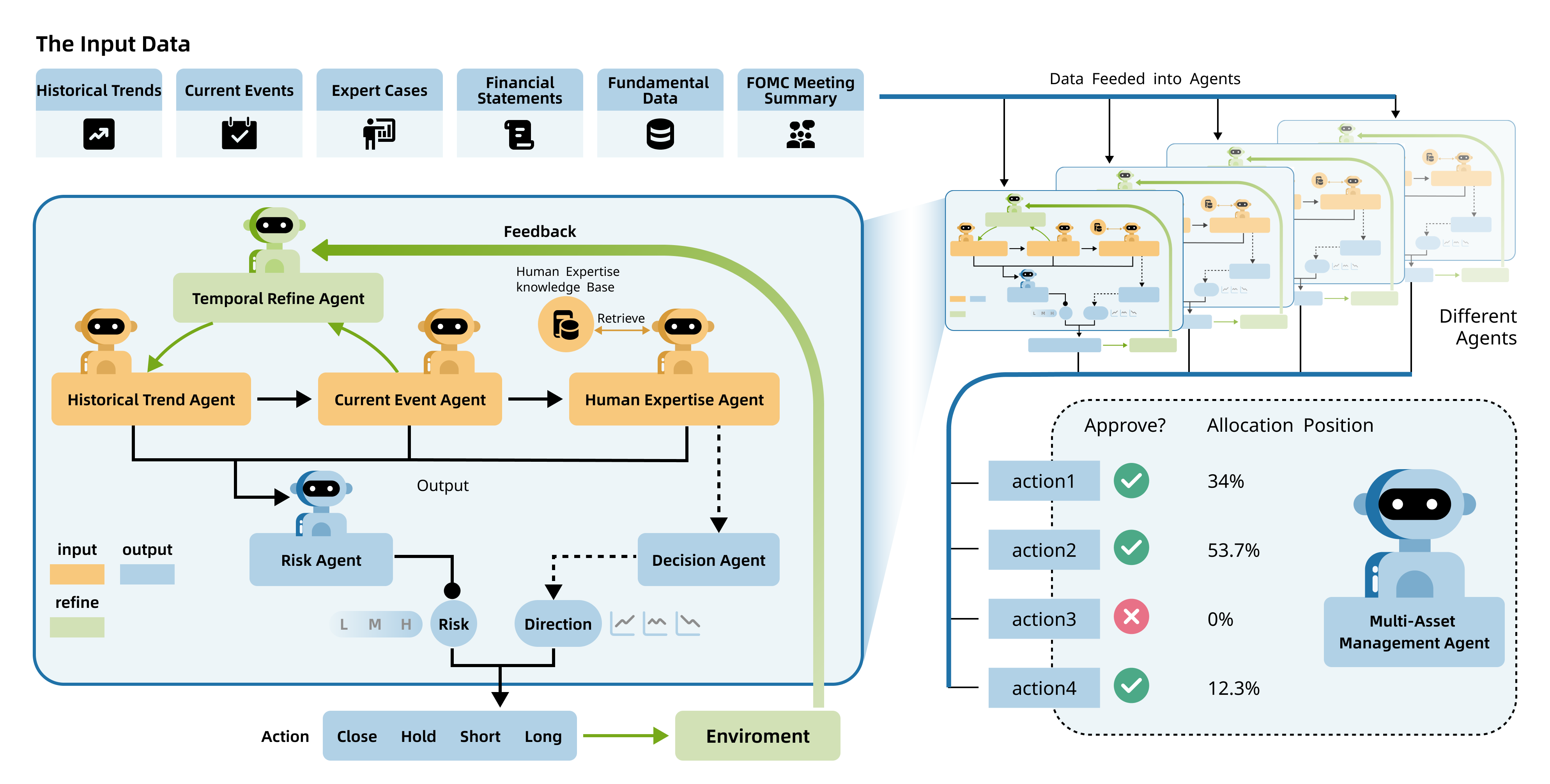}
    \caption{FinHEAR architecture. The system coordinates agents for historical trend analysis, event interpretation, and expert retrieval. These agents are organized in a temporal pipeline with feedback that adjusts past analyses based on outcomes. Risk and direction are predicted separately and combined to generate final trading actions. In the multi-asset setting, actions are dynamically aggregated based on risk-aware signals to determine final allocation positions.}
    \label{fig:framework}
\end{figure*}

\subsection{Human Expertise Curation}

Inspired by the success of legendary investors such as Warren Buffett, George Soros, and Peter Lynch, we integrate structured expert reasoning into FinHEAR. These investors embody diverse philosophies—from value investing to macro-driven speculation—offering robust principles for decision-making under uncertainty.

We construct a human-aligned knowledge base by curating writings from expert letters, books, and interviews, segmenting them into passages, and extracting structured entries: the referenced financial event, the expert’s rationale, and contextual attributes (e.g., market condition, asset type, outcome). This enables FinHEAR to retrieve analogous cases and ground reasoning in interpretable, expert-informed insights.

\paragraph{Structured Knowledge Construction.}
Using task-specific prompts and the \textbf{GPT-4o} API, we transform the curated corpus into structured \emph{query–knowledge} pairs. The resulting repository bridges unstructured narratives with actionable decision contexts for downstream reasoning tasks.

\paragraph{Verification and Adaptive Refinement.}
During inference, real-time financial news is converted into conceptual queries, embedded via \textbf{text-embedding-3-large}, and matched to expert queries using cosine similarity. Relevant knowledge activates the \textbf{Human Expertise Agent} to guide predictions. Successful predictions reinforce the retrieved knowledge, while failures trigger an adaptive refinement loop via the \textbf{Temporal Refine Agent}, which updates expert entries using GPT-4o. This iterative process ensures the knowledge base remains reliable and adaptive to evolving market conditions.

\subsection{Architecture of FinHEAR}

FinHEAR is a multi-agent reasoning framework for financial decision-making, designed to incorporate structured human expertise, adaptive risk modeling, and feedback-driven temporal refinement. As illustrated in Figure~\ref{fig:framework}, it orchestrates six specialized agents, each aligned with a key component of the reasoning process: capturing historical trends, interpreting current events, retrieving expert analogies, estimating risk, generating trading decisions, and refining predictions based on feedback. These components are grounded in behavioral economic principles such as bounded rationality, information asymmetry, and loss aversion. Further theoretical foundations are detailed in Appendix~\ref{app:bef}.

\textbf{Historical Trend Agent.}
This agent summarizes sequential market patterns across multiple time horizons, capturing price dynamics, volatility shifts, and cyclical behaviors. These patterns establish the historical context for interpreting new signals. Importantly, the trend representations are dynamically updated using realized outcomes, preventing error accumulation from earlier misinterpretations.

\textbf{Current Event Agent.}
This agent interprets real-time financial events—including macroeconomic announcements and firm-specific news—and evaluates their directional influence relative to the evolving market context. The analysis considers both short-term impacts and structural long-term implications, ensuring temporal coherence with the historical trend agent.

\textbf{Human Expertise Agent.}
To incorporate analogical reasoning from experienced investors, we construct a curated expert knowledge base. Given a current market state $x_{\text{current}}$, the agent retrieves the most relevant expert case $x_{\text{exp}}$ using event context embeddings. Each context is encoded via a transformer-based encoder $\phi(\cdot)$, and similarity is computed as:
\begin{equation}
x_{\text{retrieved}} = \arg\max_{x_{\text{exp}} \in \mathcal{M}} \cos\left( \phi(x_{\text{current}}), \phi(x_{\text{exp}}) \right)
\end{equation}
where $\mathcal{M}$ is the expert memory base. The retrieved case informs high-level decision reasoning based on past expert behavior.

\textbf{Risk Analysis Agent.}
This agent evaluates disagreement across the historical, event, and expert agents to estimate overall uncertainty. It outputs a discrete risk level—\texttt{low}, \texttt{medium}, or \texttt{high}—used later to adapt trading actions and the detailed rationale behind it.

\textbf{Decision Agent.}
The Decision Agent predicts market direction $d \in {\texttt{up}, \texttt{down}}$ using integrated historical, event, and expert signals. This guides the downstream module in selecting trading actions (e.g., \texttt{long}, \texttt{short}, \texttt{hold}, or \texttt{close}) based on both the predicted trend and calibrated risk-aware position scores.

\textbf{Temporal Refinement Agent.}
This agent strengthens temporal reasoning by refining past analyses in light of realized outcomes. When predictions conflict with actual market behavior, this agent updates historical trends to correct errors and ensure consistency across time.

\subsection{Adaptive Risk-Aware Decision Making}

\textbf{Single-Asset Setting.}
Following Prospect Theory, we distinguish three \textit{risk contexts} based on inter-agent coordination and estimated market feedback, and sample a sensitivity score $\rho$ to modulate decision strength. Rather than using uniform distributions, we adopt scaled Beta distributions with asymmetric shapes to reflect context-specific risk preferences:
\begin{equation}
\rho \sim \text{ScaledBeta}(\alpha_r, \beta_r; [a_r, b_r])
\label{eq:rho_sampling}
\end{equation}
In \texttt{low-risk contexts}—often corresponding to potential gain regions—$\rho$ follows a left-skewed distribution (e.g., $\text{Beta}(5,2)$ on $[0.75, 0.9]$), encouraging moderate risk-taking. In contrast, \texttt{high-risk contexts} adopt a right-skewed shape (e.g., $\text{Beta}(2,5)$ on $[0.1, 0.4]$), amplifying caution. This design mimics behavioral asymmetries observed in decision-making under uncertainty.

The final position score $w$ is computed by combining risk sensitivity, expert reliability, and event similarity:
\begin{equation}
w = F(\rho, \alpha, \gamma) = f(\rho) \cdot h(\alpha) \cdot g(\gamma)
\label{eq:position_score}
\end{equation}
Here, $f(\rho)$ modulates decision strength, which amplifyies actions under low risk and attenuating them under high risk, in line with Prospect Theory’s asymmetric treatment of gains and losses. Expert reliability is given by $h(\alpha) = \max(\alpha, \epsilon_\alpha)$, enforcing a minimum trust threshold under weak signals. For event similarity, $g(\gamma) = \sqrt{\max(\gamma, \epsilon_\gamma)}$ expands low values to avoid over-penalizing partially relevant historical cases. Together, these components stabilize decision outputs under uncertainty and improve robustness in noisy environments.

To select an action, we evaluate the predicted direction $d \in \{\texttt{up}, \texttt{down}\}$ together with the computed position score $w$. We introduce two thresholds, $\delta_{\text{low}}$ and $\delta_{\text{high}}$, to regulate decision-making: the lower threshold suppresses weak or ambiguous signals, while the upper threshold prevents overconfident actions under high risk. When $w$ falls below $\delta_{\text{low}}$, the agent chooses to \texttt{hold}; if $w$ exceeds $\delta_{\text{high}}$, it chooses to \texttt{close}. For intermediate scores, directional context determines the final action—\texttt{long} if the market is trending upward, or \texttt{short} if trending downward:
\begin{equation}
a = 
\begin{cases}
\texttt{close}, & w > \delta_{\text{high}} \\
\texttt{hold}, & w < \delta_{\text{low}} \\
\texttt{long}, & d = \texttt{up} \text{ and } \delta_{\text{low}} \leq w \leq \delta_{\text{high}} \\
\texttt{short}, & d = \texttt{down} \text{ and } \delta_{\text{low}} \leq w \leq \delta_{\text{high}}
\end{cases}
\label{eq:action_single}
\end{equation}

\textbf{Multi-Asset Setting.}
For $N$ assets, each stock $i$ computes a confidence score $s_i$ as in Equation~\ref{eq:position_score}, then applies a temperature-scaled softmax:

\begin{equation}
w_i = \frac{\exp(s_i / T)}{\sum_{j=1}^N \exp(s_j / T)}.
\label{eq:softmax_multi}
\end{equation}
Here the temperature $T$ adjusts allocation sharpness: lower $T$ sharpens focus, higher $T$ promotes diversification. This generalizes risk-aware decision-making to portfolio-level allocation.

\section{Experiments}
  
In this section, we empirically evaluate the performance of \textbf{FinHEAR} by addressing the following research questions:

\begin{compactitem}
    \item \textbf{RQ1: Forecasting Accuracy. } How accurately can FinHEAR predict the directional trend of major financial assets?
    
    \item \textbf{RQ2: Backtest Performance. } How well does FinHEAR perform in historical market simulations, in terms of return and risk?
    
    \item \textbf{RQ3: Component Contribution. } What is the individual impact of each module within FinHEAR on the overall trading outcome?
    
    \item \textbf{RQ4: Asset-Style Adaptability. } How does FinHEAR's performance differ for specific assets when guided by human knowledge agents with varying investment philosophies?
\end{compactitem}

\subsection{Experimental Setup}
\textbf{Datasets.\;}
We primarily focus on three stocks from different sectors: Apple Inc. (AAPL), Tesla, Inc. (TSLA), and Exxon Mobil Corporation (XOM), as well as one commodity asset: Gold (XAUUSD). For these assets, we construct a multi-source financial dataset spanning December 2019 to December 2023, comprising daily prices (open, high, low, close, volume), company financial statements, daily public news and important macroeconomic literature and data such as the official minutes of the U.S. Federal Open Market Committee (FOMC) and periodic reports of the Consumer Price Index (CPI). Each data type is routed to specialized analytical agents based on its temporal and structural characteristics, as detailed in Appendix ~\ref{app:dataset}. 

To incorporate long-term investment insights, we compile a corpus of “investor wisdom”—including books, interviews, and historical trading records of iconic investors (e.g., Buffett, Soros)—and dispatch asset-specific queries to retrieval agents aligned with each investor's philosophy. Further details about human expertise base are available in Appendix ~\ref{app:human expertise}.

\begin{table*}[!t]

\caption{Comparison of the forecasting trend task of FinHEAR and the other 9 baselines in Single-Asset during the test period}
\resizebox{\linewidth}{!}{
\begin{tabular}{@{\extracolsep{\fill}}llcccccccc@{}}
\toprule
\textbf{Category} & \textbf{Method} 
& \multicolumn{2}{c}{\textbf{XAUUSD}} 
& \multicolumn{2}{c}{\textbf{AAPL}} 
& \multicolumn{2}{c}{\textbf{TSLA}} 
& \multicolumn{2}{c}{\textbf{XOM}} \\
\cmidrule(r){3-4} \cmidrule(r){5-6} \cmidrule(r){7-8} \cmidrule(r){9-10}
& & \textbf{ACC (\%)}~$\uparrow$ & \textbf{MCC}~$\uparrow$ 
  & \textbf{ACC (\%)}~$\uparrow$ & \textbf{MCC}~$\uparrow$ 
  & \textbf{ACC (\%)}~$\uparrow$ & \textbf{MCC}~$\uparrow$
  & \textbf{ACC (\%)}~$\uparrow$ & \textbf{MCC}~$\uparrow$ \\
\midrule
\multirow{2}{*}{\centering Rule-based} 
    & Momentum        & 47.58 & -0.049  & 48.58 & -0.0365  & 51.82 & 0.0257  & 47.18 & -0.0565 \\
    & Mean Reversion  & 47.35 & -0.053  & 45.49 & -0.0742 & 47.76 & -0.0372 & 52.65 & 0.0521 \\
\midrule
\multirow{4}{*}{\centering Deep-learning} 
    & ALSTM           & 54.32 & 0.1212  & 54.69 & 0.1739  & 54.88 & 0.1095  & 52.25 & 0.0591 \\
    & Informer        & 53.70 & 0.1576  & 53.52 & 0.1526  & 55.36 & 0.0523  & 53.59 & 0.0836 \\
    & StockNet        & 52.57 & -0.0366 & 53.59 & 0.0872  & 55.77 & 0.0416  & 51.96 & 0.0493 \\
    & PPO             & 54.72 & 0.1346  & 53.62 & 0.1815  & 54.07 & -0.0094 & 56.52 & 0.1237 \\
\midrule
\multirow{4}{*}{\centering LLM-based} 
    & CoT             & 52.03 & 0.1493  & 56.13 & 0.1629  & 51.83 & 0.0623  & 51.27 & 0.0585 \\
    & GA              & 54.47 & 0.1551  & 54.17 & 0.1082  & 56.54 & 0.1412  & 52.78 & 0.0624 \\
    & FinMem          & 55.73 & 0.1423  & 55.32 & 0.1327  & 58.82 & 0.1328  & 53.91 & 0.0430 \\
    & FinCon          & 56.81 & 0.1408  & 52.76 & 0.0709  & 57.46 & 0.1194  & 56.37 & 0.1428 \\
\midrule
\rowcolor{gray!20} 
\multirow{1}{*}{\centering Ours} 
    & \textbf{FinHEAR} & \textbf{58.97} & \textbf{0.2017}  & \textbf{59.08} & \textbf{0.1964}  & \textbf{61.56} & \textbf{0.1897}  & \textbf{58.19} & \textbf{0.1625} \\
\bottomrule
\end{tabular}
}

\label{tab:forecasting}
\end{table*}
\textbf{Baselines.\;}
To ensure a thorough evaluation across diverse methodological paradigms, we adopt  three categories of baselines:
\begin{compactenum}
\item
\textbf{Rule-based methods}: including momentum and mean-reversion strategies, adhering to predefined financial trading rules. 
\item 
\textbf{Deep learning methods}: including ALSTM, Informer, StockNet, DQN and PPO, representing widely used neural approaches for sequential and decision-making tasks. 

\item 

\textbf{LLM-based agents}: including general-purpose generative models(GA), chain-of-thought reasoning approaches(CoT), FinMem and FinCon, covering a broad spectrum of large language model applications in finance. 
\end{compactenum}


\textbf{Evaluation Metrics. \;}
We evaluate performance under two complementary settings: \
(1) \textbf{Trend Forecasting}, measured by Accuracy (ACC\%) and Matthews Correlation Coefficient (MCC), which assess the correctness and robustness of next-day market predictions. \
(2) \textbf{Decision-making}, measured by Cumulative Return (CR\%)~\cite{hull2012risk}, Sharpe Ratio (SR)~\cite{sharpe1994sharpe}, Maximum Drawdown (MDD\%)~\cite{ang2006downside} and Calmar Ratio(Calmar R)~\cite{young1991calmar}, which quantify profitability, risk-adjusted return, and resilience under adverse conditions. \
Formal definitions and formulas are provided in the Appendix~\ref{app:metrics}.

\textbf{Implementation Details. \;}
To ensure fair comparison, all LLM-based methods, including FinHEAR, use GPT-4o-mini as the underlying model. The data was split into a training period (January 2, 2020 - December 29, 2022) and a test period (January 2, 2023 - December 29, 2023). Both Deep Learning and LLM-based methods were trained on the designated training data and evaluated over the specified test period. Detailed information regarding the data split and evaluation periods is provided in Appendix~\ref{app:datasplit}. The detailed experimental configurations for Trend Forecasting and Trading Decision-making are provided in Appendix~\ref{app:forecasting config} and Appendix~\ref{app:trading config}, respectively. In the main comparative experiments, all models uniformly adopted a fixed 5-day input time window. This setting was maintained consistently throughout subsequent experiments.

\subsection{Experimental Results for Trend Forecasting Task}

Despite the potential decoupling between predictive accuracy and financial returns~\cite{de2018advances}, we argue that accurate trend forecasting is fundamental to the success of financial trading strategies, as predictive quality directly determines directional decisions. Hence, our initial focus is on financial trend prediction as the foundation for strategy development. Specifically, to answer RQ1, we evaluate FinHEAR’s performance in predicting next-day price movements of financial assets. 
 
Table~\ref{tab:forecasting} reports trend forecasting results across three datasets. FinHEAR consistently achieves the highest Accuracy (ACC) and Matthews Correlation Coefficient (MCC)~\cite{matthews1975comparison}, demonstrating strong robustness in single-asset prediction. Overall, rule-based methods exhibit unstable performance due to the rigidity of their rules. Deep learning methods demonstrated relatively stable performance during the subsequent test period. In contrast, LLM-based methods exhibited a wider performance range. This is because their predictions rely on complex strategies or reasoning. While this approach enables potential for high returns, it also renders these methods vulnerable to market changes.

FinHEAR addresses these limitations through two key innovations: expert-guided reasoning to reduce information asymmetry, and feedback-driven refinement to support adaptive temporal decision-making. Its asset-specialized agent design further enhances robustness and interpretability across diverse and volatile market conditions.

\begin{table*}[!t]
\caption{Comparison of the overall trading performance of FinHEAR and the other 9 baselines in Single-Asset during the test period}

\centering
\resizebox{\linewidth}{!}{
\begin{tabular}{@{\extracolsep{\fill}}llcccc|cccc|cccc|cccc@{}}
\toprule
\textbf{Category} & \textbf{Method} 
& \multicolumn{4}{c|}{\textbf{XAUUSD}} 
& \multicolumn{4}{c|}{\textbf{AAPL}} 
& \multicolumn{4}{c|}{\textbf{TSLA}} 
& \multicolumn{4}{c}{\textbf{XOM}} \\
\cmidrule(r){3-6} \cmidrule(r){7-10} \cmidrule(r){11-14} \cmidrule(r){15-18}
& & \textbf{CR (\%)}~$\uparrow$ & \textbf{SR}~$\uparrow$ & \textbf{MDD (\%)}~$\downarrow$ & \textbf{CalmarR}~$\uparrow$ 
  & \textbf{CR (\%)}~$\uparrow$ & \textbf{SR}~$\uparrow$ & \textbf{MDD (\%)}~$\downarrow$ & \textbf{CalmarR}~$\uparrow$ 
  & \textbf{CR (\%)}~$\uparrow$ & \textbf{SR}~$\uparrow$ & \textbf{MDD (\%)}~$\downarrow$ & \textbf{CalmarR}~$\uparrow$ 
  & \textbf{CR (\%)}~$\uparrow$ & \textbf{SR}~$\uparrow$ & \textbf{MDD (\%)}~$\downarrow$ & \textbf{CalmarR}~$\uparrow$ \\
\midrule
\multirow{2}{*}{Rule-based} 
 & Momentum & -0.84 &  -0.28 & 13.46 & -0.06 & -4.04 & -0.31 & 19.87 & -0.20 & 40.73 & 0.85 & 25.58 & 1.58 & -31.32 & -1.56 & 36.57 & -0.86 \\
 & Mean Reversion & -19.00 & -1.80 & 19.82 & -0.96 & -12.15 & -0.78 & 19.78 & -0.61 & -42.70 & -0.89 & 62.07 & -0.69 & 25.09 & 0.89 & \textbf{12.90} & \textbf{1.94} \\
\midrule
\multirow{4}{*}{Deep-learning} 
 & ALSTM & 10.45 & 0.53 & 14.86 & 0.70 & 16.77 & 0.75 & 21.49 & 0.78 & 11.47 & 1.14 & \textbf{23.71} & 0.48 & -0.78 & -0.36 & 18.45 & -0.04 \\
 & Informer & 5.36 & 1.33 & 17.34 & 0.31 & 8.54 & 1.22 & 17.34 & 0.49 & -5.42 & -1.22 & 25.12 & -0.22 & -14.14 & -2.19 & 18.84 & -0.75 \\
 & StockNet & -19.78 & -1.73 & 23.90 & -0.83 & 12.36 & 0.85 & 20.83 & 0.59 & 8.97 & 0.42 & 19.97 & 0.45 & -5.29 & -0.33 & 27.91 & -0.19 \\
 & PPO & 6.82 & 1.77 & \textbf{15.85} & 0.43 & 15.46 & 1.67 & 19.35 & 0.80 & -2.82 & -0.01 & 25.11 & -0.11 & 13.10 & 1.97 & 25.93 & 0.51 \\
\midrule
\multirow{4}{*}{LLM-based} 
 & CoT & 13.98 & 1.01 & 18.06 & 0.77 & 22.07 & 1.32 & 25.46 & 0.87 & -14.26 & -1.52 & 46.61 & -0.86 & -9.83 & -0.51 & 31.18 & -0.32 \\
 & GA & 14.51 & 1.78 & 25.37 & 0.57 & 13.66 & 1.32 & \textbf{15.02} & 0.91 & 16.98 & 0.76 & 33.98 & 0.50 & -15.59 & -1.24 & 33.74 & -0.46 \\
 & FinMem & 18.32 & 1.52 & 18.12 & 1.01 & 19.24 & 1.43 & 31.02 & 0.62 & 28.85 & 1.22 & 33.84 & 0.85 & -20.15 & -1.59 & 38.59 & -0.52 \\
 & FinCon & 17.29 & 1.48 & 19.25 & 0.90 & -8.17 & -0.53 & 32.47 & -0.36 & 33.73 & 1.47 & 29.17 & 1.16 & 23.95 & 1.77 & 31.72 & 0.76 \\
\midrule
\rowcolor{gray!10}
\multirow{1}{*}{Ours} 
 & \textbf{FinHEAR} & \textbf{32.45} & \textbf{1.84} & 18.47 & \textbf{1.14} & \textbf{30.81} & \textbf{1.95} & 23.88 & \textbf{1.29} & \textbf{58.74} & \textbf{1.62} & 25.21 & \textbf{2.33} & \textbf{37.16} & \textbf{2.25} & 29.61 & 1.25 \\
\bottomrule
\end{tabular}
}

\label{tab:performance}
\end{table*}
\subsection{Experimental Results for Trading Decision-making Task}

\paragraph{Single-Asset Trading Task}

To evaluate the trading performance of FinHEAR, we compare it against rule-based strategies, deep learning models, and LLM-based agents across multiple dimensions on four financial assets. As shown in Table~\ref{tab:performance}, FinHEAR consistently achieves the highest Cumulative Return and Sharpe Ratio, indicating superior performance in both return generation and stability. Further details of these experimental results are provided in Appendix~\ref{app:singleasset}.This advantage holds across diverse asset types, thereby showcasing the framework's generalizability under varying market dynamics.

FinHEAR addresses these limitations through the integration of expert knowledge to guide decisions, the adjustment of position sizes based on risk perception and behavioral traits like loss aversion, and the refinement of past predictions through feedback. These capabilities enable FinHEAR to adapt effectively in dynamic markets and deliver robust, resilient trading performance.

We further conducted a comprehensive backtest incorporating conservative transaction costs (5 basis points per side). As detailed in Appendix~\ref{app:transaction_costs}, our findings remain robust, with \textsc{FinHEAR} consistently outperforming strong baselines even after accounting for realistic trading frictions.
\paragraph{Robustness Across Market Regimes}
We evaluate FinHEAR on XAUUSD in 2023 across two phases: a risk-aversion uptrend and a rate-hike downtrend. As shown in \textbf{Table~\ref{tab:market_conditions}}, FinHEAR achieves the highest returns and Sharpe ratios in both regimes and uniquely remains profitable under sustained bearish conditions.

\begin{table}[h]
\centering
\small
\setlength{\tabcolsep}{5pt} 
\renewcommand{\arraystretch}{1.15} 
\begin{tabular}{llccc}
\toprule
\textbf{Market Condition} & \textbf{Methods} & \textbf{CR} & \textbf{SR} \\
\midrule
\multirow{6}{*}{\parbox{3.5cm}{\centering Risk-Aversion Driven Uptrend \\ \small(Feb 1 -- May 4, 2023)}} 
    & FinCon  & 14.38 & 2.30 \\
    & PPO     & 2.72  & 1.02 \\
    & ALSTM   & 15.56 & 4.26 \\
    & CoT     & 19.38 & 3.63 \\
    & FinMem  & 12.82 & 2.45 \\
    & FinHEAR & \textbf{18.25} & \textbf{4.79} \\
\midrule
\multirow{6}{*}{\parbox{3.5cm}{\centering Rate-Hike Expectation Driven Downtrend \\ \small(May 5 -- Oct 5, 2023)}}
    & FinCon  & -5.58  & -0.14 \\
    & PPO     & -3.83  & -0.06 \\
    & ALSTM   & -10.39 & -2.51 \\
    & CoT     & -10.12 & -2.14 \\
    & FinMem  & -8.98  & -2.27 \\
    & FinHEAR & \textbf{7.82} & \textbf{0.47} \\
\bottomrule
\end{tabular}
\caption{Performance comparison under different market conditions. CR = cumulative return, SR = Sharpe ratio.}
\label{tab:market_conditions}
\end{table}
\paragraph{Multi-Asset Portfolio Management Task}

To evaluate FinHEAR's capabilities in portfolio management, we extend its multi-agent structure (previously designed for single-asset trading) into a higher-level decision framework capable of handling multiple financial assets simultaneously. This extension enables FinHEAR to coordinate asset-specific reasoning while optimizing overall portfolio performance.
\begin{figure}[!htbp]

\centering
\begin{minipage}{0.48\textwidth}
    \centering
    \small
\captionof{table}{Portfolio Management Performance Evaluation: FinHEAR vs. Baselines (Test Period)}
    
    \renewcommand{\arraystretch}{1.0}
    \setlength{\tabcolsep}{4pt}
    \begin{tabular}{@{}lcccc@{}}
    \toprule
    \textbf{Strategy} & \textbf{CR (\%)}~$\uparrow$ & \textbf{SR}~$\uparrow$ & \textbf{MDD (\%)}~$\downarrow$ & \textbf{CalmarR}~$\uparrow$ \\
    \midrule
    Markowitz & 39.21 & 2.75 & \textbf{12.84} & 3.05 \\
    DQN       & 45.13 & 1.60 & 32.77         & 1.38 \\
    FinCon    & 55.02 & 2.11 & 35.15         & 1.57 \\
    \rowcolor{gray!20} 
    FinHEAR   & \textbf{83.89} & \textbf{3.28} & 15.77 & \textbf{5.32} \\
    \bottomrule
    \end{tabular}
        
        \label{tab:multi_asset_performance}
\end{minipage}
\hfill
\begin{minipage}{0.48\textwidth}
    \centering
    \includegraphics[width=\textwidth]{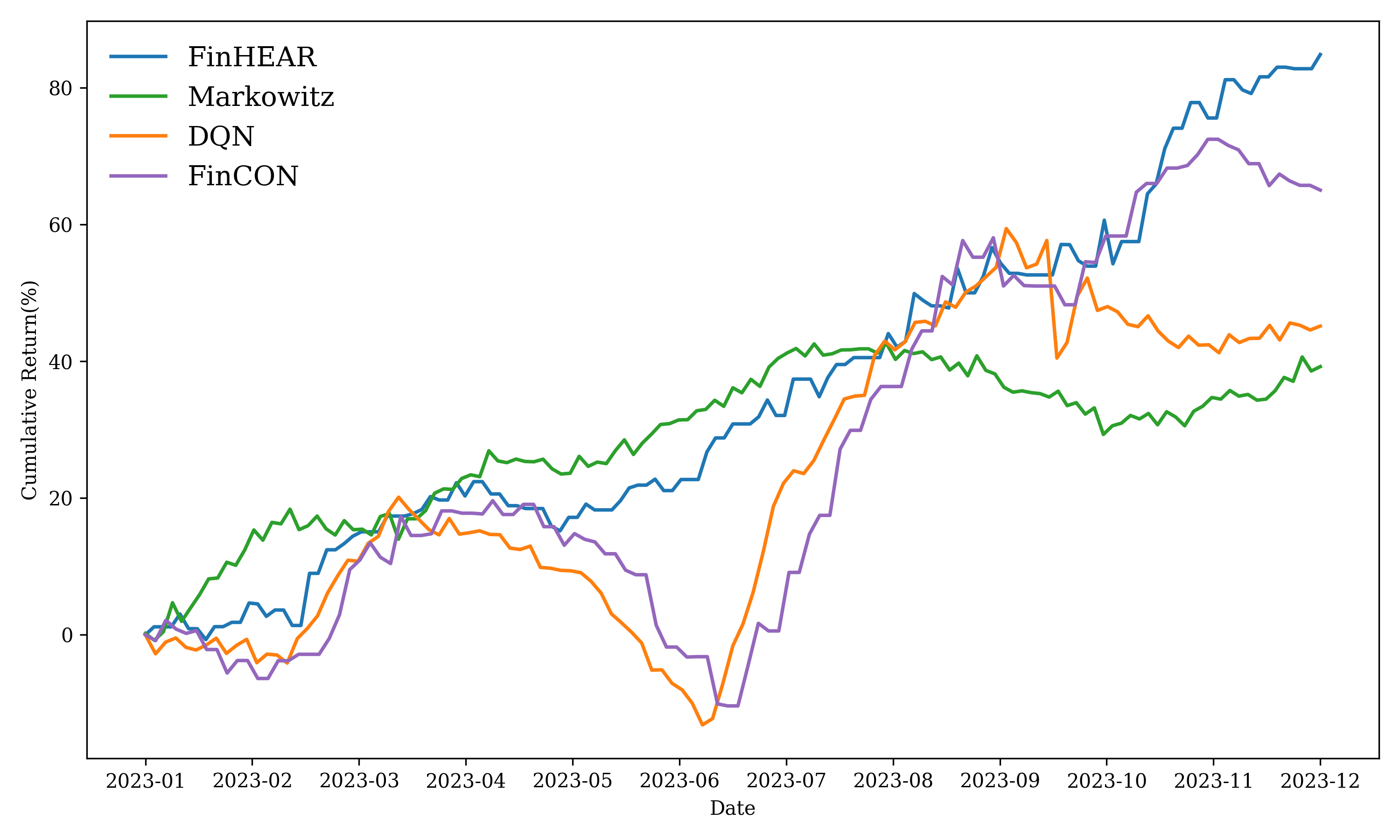}
    \caption{Multi-asset portfolio cumulative returns over time for all strategies}
    \label{fig:portfolio}
\end{minipage}
\end{figure}
We compare FinHEAR with three baselines: (1) the classical Markowitz mean-variance (MV) portfolio, where the covariance matrix and expected returns are estimated from training data~\cite{markowitz2008portfolio}; (2) a deep reinforcement learning approach (DQN)~\cite{mnih2013playing}, trained on data from three years prior to the test period, and employing a continuous action space to reflect real-world trading granularity; and (3) FinCon, a recent LLM-based agent. All methods share the same training and evaluation periods for fairness.
\begin{table}[h]
\centering
\scriptsize
\setlength{\tabcolsep}{3.8pt} 
\renewcommand{\arraystretch}{1.15} 
\begin{tabular}{lccc}
\toprule
\multirow{2}{*}{\textbf{Method}} &
\multicolumn{1}{c}{\textbf{XAUUSD}} &
\multicolumn{1}{c}{\textbf{AAPL}} &
\multicolumn{1}{c}{\textbf{TSLA}} \\
\cmidrule(lr){2-4}
 & \textbf{CR / SR} & \textbf{CR / SR} & \textbf{CR / SR} \\
\midrule
FinHEAR (Full, Deep KB) & 32.45 / 1.84 & 30.81 / 1.95 & 58.74 / 1.62 \\
FinHEAR (Simplified KB) & 24.33 / 1.51 & 19.55 / 1.32 & 39.26 / 1.51 \\
FinCon & 17.29 / 1.48 & -8.17 / -0.53 & 33.73 / 1.47 \\
PPO & 6.82 / 1.77 & 15.46 / 1.67 & -2.82 / -0.01 \\
ALSTM & 10.45 / 0.53 & 16.77 / 0.75 & 11.47 / 1.14 \\
FinMem & 18.32 / 1.52 & 19.24 / 1.43 & 28.85 / 1.22 \\
CoT (Full KB) & 18.19 / 1.45 & 25.64 / 1.59 & 10.37 / 0.61 \\
CoT & 13.98 / 1.01 & 22.07 / 1.32 & -14.26 / -1.52 \\
\bottomrule
\end{tabular}
\caption{
\textbf{Disentangling architecture and knowledge contributions.} 
Both components are significant, with maximal performance when combined. 
\textbf{CR} = Cumulative Return.
}
\label{tab:disentangle_results}
\end{table}
As shown in Table~\ref{tab:multi_asset_performance} and Figure~\ref{fig:portfolio}, FinHEAR consistently outperforms all baselines in terms of Cumulative Return (CR), Sharpe Ratio (SR), and Calmar Ratio (CalmarR). These results strongly suggest
that FinHEAR’s long-short strategy not only achieves higher excess returns but also better controls extreme risks.
\begin{table*}[]

\caption{Comparison of trading performance of different variants of FinHEAR under single-asset and portfolio management}
\centering
\small
\begin{tabular}{llrrrrr}
\toprule
\textbf{Task} & \textbf{Asset} & \textbf{Ablation Variants} 
& \textbf{CR (\%)}~$\uparrow$ 
& \textbf{SR}~$\uparrow$ 
& \textbf{MDD (\%)}~$\downarrow$ 
& \textbf{CalmarR}~$\uparrow$ \\
\midrule
\multirow{4}{*}{Single-Asset} & \multirow{4}{*}{AAPL} 
  &  w/o temporal refinement          & 22.39 & 1.67 & 25.58 & 0.88 \\
  & &  w/o past trend              & 14.41 & 0.86 & 36.96 & 0.39 \\
  & &  w/o human expertise  &  19.75 & 0.32 & 35.15 & 0.56 \\
  & &  w/o adaptive risk-aware  &  16.32 & 0.74 & 31.72 & 0.51 \\
  \rowcolor{gray!20} 
  & & FinHEAR                      & \textbf{30.81} & \textbf{1.95} & \textbf{23.88} & \textbf{1.29} \\
\midrule
\multirow{4}{*}{Portfolio Mgmt.} & \multirow{4}{*}{\begin{tabular}[c]{@{}c@{}}XAUUSD,\\ AAPL, TSLA,\\XOM\end{tabular}} 
  &  w/o temporal refinement         & 52.73 & 2.17 & 37.81 & 1.39 \\
  & &  w/o past trend              & 37.29 & 1.34 & 33.65 & 1.11 \\
  & &  w/o human expertise  & 42.12 & 1.22 & 28.94 & 1.46 \\
  & &  w/o adaptive risk-aware  & 36.18 & 1.15 & 35.03 & 1.03 \\
  \rowcolor{gray!20} 
  & & FinHEAR                      & \textbf{83.89} & \textbf{3.28} & \textbf{25.77} & \textbf{3.22} \\
\bottomrule
\end{tabular}

\label{tab:finhear-abl}
\end{table*}

\subsection{Ablation Studies}

To address RQ3 and RQ4, we conduct two ablation studies under the same training and testing setup as the main experiments. The first study, summarized in Table~\ref{tab:finhear-abl}, examines the contribution of key components within FinHEAR. The second study (Table~\ref{tab:expert-style-single}) investigates the impact of different expert styles on single-asset trading. Further experimental details are provided in Appendix~\ref{app:ablation}
\subsubsection{Ablation Study on Model Components}
To assess the role of key components in FinHEAR, we conduct ablation experiments by removing four modules: historical information, human expertise, temporal refinement, and adaptive risk-aware decision-making. Disabling the risk module reduces FinHEAR to a fixed full-position strategy, lacking flexible position control. This study aims to quantify the specific contribution of each module to FinHEAR's overall performance. As shown in Table~\ref{tab:finhear-abl}, all removals cause performance drops, though to varying degrees.

\textbf{Historical Information and Temporal Refinement. }  
Removing historical information leads to the largest decline in cumulative return and Sharpe ratio, indicating that past news is crucial for capturing market trends. This reflects the idea of \textit{path dependence}—where current behavior is shaped by prior events. Similarly, removing feedback-driven temporal refinement weakens performance by preventing the model from correcting earlier errors, thereby reducing consistency.

\textbf{Human Expertise. }  
Removing expert knowledge also degrades performance, especially in single-asset tasks. This supports the concept of \textit{bounded rationality}, where investors rely on expert-informed heuristics under uncertainty. FinHEAR benefits from these inductive biases, improving judgment in volatile or data-limited conditions.

\textbf{Adaptive Risk-Aware Decision. }  
Without adaptive risk control, FinHEAR becomes less responsive to market uncertainty, resulting in larger drawdowns and lower Calmar Ratios. This supports insights from \textit{prospect theory}, which emphasizes adjusting risk based on potential losses. The risk-aware mechanism helps FinHEAR manage exposure and improve robustness.

\subsubsection{Ablation Study on Asset-Expertise Alignment}

\begin{table}[t]
\caption{Performance of expert-style agents on single-asset trading}
\centering
\scriptsize  

\setlength{\tabcolsep}{4pt}  
\begin{tabular}{llcccc}
\toprule
\textbf{Asset} & \textbf{Expert} 
& \textbf{CR (\%)}~$\uparrow$ 
& \textbf{SR}~$\uparrow$ 
& \textbf{MDD (\%)}~$\downarrow$ 
& \textbf{CalmarR}~$\uparrow$ \\
\midrule

\multirow{4}{*}{XAUUSD} 
    
    & \cellcolor{gray!10}\textbf{Soros}   & \cellcolor{gray!10}\textbf{32.45} & \cellcolor{gray!10}\textbf{1.84} & \cellcolor{gray!10}\textbf{28.47} & \cellcolor{gray!10}\textbf{1.34} \\
    & Buffet & 25.73 & 1.52 & 31.20 & 0.82 \\
    & Lynch   & 28.98 & 1.63 & 34.15 & 0.85 \\
    & Graham  & 21.39 & 1.33 & 30.76 & 0.69 \\
\midrule
\multirow{4}{*}{AAPL} 
    & Soros   & 17.35 & 0.68 & 34.61 & 0.50 \\
    & \cellcolor{gray!10}\textbf{Buffet} & \cellcolor{gray!10}\textbf{30.81} & \cellcolor{gray!10}1.95 &\cellcolor{gray!10} \textbf{23.88} & \cellcolor{gray!10}\textbf{1.29} \\
    & Lynch   & 26.42 & 1.57 & 28.95 & 0.91 \\
    & Graham  & 28.71 & \textbf{1.98} & 25.16 & 1.14 \\
\midrule
\multirow{4}{*}{TSLA} 
    & Soros   & 21.26 & 1.19 & 38.30 & 0.55 \\
    & Buffet & 33.75 & 1.28 & \textbf{32.11} & 1.02 \\
    & \cellcolor{gray!10}\textbf{Lynch}   &\cellcolor{gray!10} \textbf{48.74} &\cellcolor{gray!10} \textbf{1.62} &\cellcolor{gray!10} 35.21 &\cellcolor{gray!10} \textbf{1.38} \\
    & Graham  & 35.52 & 1.49 & 39.18 & 1.10 \\
\midrule
\multirow{4}{*}{XOM} 
    & Soros   & 25.19 & 1.54 & 37.12 & 0.68 \\
    & Buffet & 21.62 & 1.36 & 25.51 & 0.85 \\
    & Lynch   & 15.38 & 0.88 & 44.01 & 0.35 \\
    & \cellcolor{gray!10}\textbf{Graham}  &\cellcolor{gray!10} \textbf{27.16} &\cellcolor{gray!10} \textbf{2.25} &\cellcolor{gray!10} \textbf{29.61} &\cellcolor{gray!10} \textbf{0.91} \\
\bottomrule
\end{tabular}

\label{tab:expert-style-single}
\end{table}

In our second study, we examine whether different expert investment styles improve single-asset predictions by aligning better with specific asset characteristics. This setup reflects key concepts from behavioral economics, where bounded rationality and information asymmetry lead investors to adopt simplified heuristics based on personal or institutional experience. Table~\ref{tab:expert-style-single} shows clear style-asset alignment such as Buffett-style on AAPL. These results suggest that investment philosophies vary in their effectiveness across different market conditions, and encoding such distinctions can enhance model alignment with asset-specific dynamics.

In addition to investigating expert styles, we also examined how model performance is affected by varying input time window sizes. This study revealed that optimal prediction time windows differ across assets, exhibiting \textbf{Temporal Specificity of Asset Performance}. Detailed experimental results and analysis concerning different time windows are provided in Appendix~\ref{app:ablation}.

Overall, these findings underscore FinHEAR's flexibility in capturing behaviorally grounded strategies through expert-guided specialization. By embedding decision principles such as bounded rationality and loss aversion, FinHEAR supports more interpretable and human-aligned reasoning—positioning it as a behavior-aware framework for expert-informed decision-making in real-world financial contexts.

\subsubsection{Disentangling Architecture vs. Knowledge Base Contributions}
We perform a two-path ablation to isolate contributions from architecture and knowledge:

\begin{itemize}
    \item \textbf{FinHEAR (Simplified KB):} Full architecture with a simplified, keyword-only knowledge base.
    \item \textbf{CoT (Full KB):} Standard Chain-of-Thought baseline augmented with our curated knowledge base.
\end{itemize}

\textbf{Table~\ref{tab:disentangle_results}} shows that:  
(1) FinHEAR’s architecture improves performance even with minimal knowledge;  
(2) the curated knowledge base brings substantial independent gains;  
(3) combining both achieves the best results, confirming their synergy.

\subsubsection{Sensitivity to Risk Thresholds}
To validate our implementation of Prospect Theory, we conduct a sensitivity analysis on the asymmetric risk thresholds $\delta_{\text{low}}$ and $\delta_{\text{high}}$ used in our Risk Agent. 

\begin{table}[h]
\centering
\scriptsize
\setlength{\tabcolsep}{4pt} 
\renewcommand{\arraystretch}{1.15} 
\begin{tabular}{lcccc}
\toprule
 & \multicolumn{4}{c}{$\delta_{\text{high}}$} \\
\cmidrule(lr){2-5}
$\delta_{\text{low}}$ & \textbf{0.5} & \textbf{0.6} & \textbf{0.7} & \textbf{0.8} \\
\midrule
0.1 & 26.88 / 1.71 & 28.13 / 1.83 & 29.5 / 1.75  & 28.05 / 1.48 \\
0.2 & 29.03 / 1.81 & 30.81 / 1.95 & 28.03 / 1.83 & 30.24 / 1.53 \\
0.3 & 25.66 / 1.67 & 25.93 / 1.63 & 27.55 / 1.92 & 28.15 / 1.91 \\
0.4 & 15.34 / 1.23 & 22.53 / 1.51 & 24.88 / 1.79 & 27.89 / 1.99 \\
0.5 & --           & 20.37 / 1.14 & 21.05 / 1.52 & 24.11 / 1.36 \\
0.6 & --           & --           & 16.97 / 1.39 & 18.71 / 1.23 \\
0.7 & --           & --           & --           & 7.36 / 1.24  \\
\bottomrule
\end{tabular}
\caption{Impact of $(\delta_{\text{low}}, \delta_{\text{high}})$ thresholds on cumulative return (\textbf{CR}) and Sharpe ratio (\textbf{SR}). Each entry shows \textbf{CR/SR}.}
\label{tab:delta_ablation}
\end{table}
As shown in \textbf{Table~\ref{tab:delta_ablation}}, the optimal thresholds $(\delta_{\text{low}}=0.2, \delta_{\text{high}}=0.6)$ align with Prospect Theory's prediction: agents exhibit stronger risk aversion in loss domains while remaining moderately risk-seeking in gain domains.

\section{Conclusion}
This paper has proposed FinHEAR to address three key challenges in financial decision-making: incomplete information, risk sensitivity, and temporal consistency. To reduce information asymmetry, it retrieves expert cases to emulate bounded rationality. To manage risk, it adjusts positions based on prediction uncertainty. A feedback mechanism further refines past decisions using outcome-driven updates.
These components form a structured, event-driven pipeline grounded in behavioral finance. Experiments on single-asset and portfolio tasks show that FinHEAR consistently outperforms rule-based, deep learning, and LLM-based baselines across forecasting and decision-making metrics.
Overall, FinHEAR offers a promising step toward interpretable, resilient, and behavior-aware financial decision-making.

\section*{Limitations}

While FinHEAR incorporates core principles from behavioral economics, the integration between AI reasoning and human cognitive theories remains preliminary. Future work could explore more intelligent and adaptive selection of expert personas based on asset context, investor profiles, or market conditions, enabling deeper alignment between agent strategies and human decision-making behavior.

\section*{Acknowledgments}

\bibliography{references}

\newpage
\appendix

\appendix
\section{Appendix}
\subsection{Behavioral Economics Foundations}
\label{app:bef}
FinHEAR is grounded in key principles from behavioral economics, which guide its architectural design and decision-making mechanisms:

\begin{itemize}

    \item \textbf{Bounded Rationality}~\cite{BoundedRationality,bounded}: Humans have limited cognitive resources and cannot process all available information. FinHEAR incorporates expert heuristics to emulate simplified, experience-driven reasoning processes under uncertainty.

    \item \textbf{Information Asymmetry}~\cite{InformationAsymmetry1,InformationAsymmetry2}: Markets often contain uneven distributions of information. By retrieving expert-informed cases and structuring external event signals, FinHEAR reduces decision gaps caused by partial information.

    \item \textbf{Prospect Theory}~\cite{ProspectTheoryloss,ProspectTheory,ProspectTheoryrisk}: Human decision-making under risk exhibits loss aversion—losses are weighed more heavily than equivalent gains. FinHEAR models this asymmetry by modulating position sizes according to risk sensitivity and prediction confidence.

    \item \textbf{Behavioral Portfolio Theory}~\cite{behavioral}: Investors mentally segment their portfolios into layers (e.g., safety vs. growth). FinHEAR reflects this layered decision logic by supporting adaptive strategies across heterogeneous asset classes.

\end{itemize}

Together, these theoretical foundations support FinHEAR's goal of aligning algorithmic behavior with real-world investor psychology, promoting interpretability and decision relevance in financial environments.

\subsection{Evaluation Metrics}
\label{app:metrics}
We employ five widely-used metrics in finance to compare the investment rewards of FinHEAR against other algorithmic trading agents. Here are their introductions:
\begin{itemize}

    \item \textbf{Cumulative Return}
    
    Cumulative return(CR) measures a strategy’s profitability over time. In our framework, it is computed by weighting the log return $\ln(p_{t+1}/p_t)$—where $p_t$ and $p_{t+1}$ are prices at time $t$ and $t{+}1$—by the trading action $a_t$ and position size $w_t$ at each step. Summing over all steps yields the total cumulative return, allowing us to quantify how each decision and exposure contribute to overall performance:
    \begin{equation}
    \small\text{CR} = \sum_{t=1}^{n} \ln\left(\frac{p_{t+1}}{p_t}\right) \cdot a_t \cdot w_t
    \end{equation}
    \item \textbf{Sharp Ratio}

    The Sharpe Ratio (SR) is a standard metric for evaluating risk-adjusted return. It measures the excess return $R_p - R_f$ relative to its volatility $\sigma_p$, where $R_p$ is the average return of the strategy and $R_f$ is the risk-free rate. A higher SR indicates better performance per unit of risk:
\begin{equation}
    \small\text{SR} = \frac{R_p - R_f}{\sigma_p}
\end{equation}
    \item \textbf{Max Drawdown}
    
    The Maximum Drawdown (MDD) quantifies the worst observed loss over a period by measuring the largest peak-to-trough drop in value. Given a peak value $P_{\text{peak}}$ and a subsequent trough $P_{\text{trough}}$, MDD reflects the largest percentage decline before recovery:

\begin{equation}
\small\text{MDD} = \max\left( \frac{P_{\text{peak}} - P_{\text{trough}}}{P_{\text{peak}}} \right)    
\end{equation}
    \item \textbf{Calmar Ratio}
    The Calmar Ratio (CalmarR) evaluates risk-adjusted return using MDD as the risk denominator. It measures how efficiently a strategy generates returns relative to its largest drawdown. Typically, $R_p$ is the annualized strategy return and $R_f$ is the annualized risk-free rate:

\begin{equation}
\small\text{CalmarR} = \frac{R_p - R_f}{\text{MDD}}    
\end{equation}
\item \textbf{Accuracy}

For evaluating our model's performance in the forecasting task, we employ Accuracy (ACC). This metric is defined as:
\begin{equation}
\small\text{ACC} = \frac{\text{TP} + \text{TN}}{\text{TP} + \text{TN} + \text{FP} + \text{FN}}     
\end{equation} 
True Positives (TP) refer to correctly predicted price increases, while True Negatives (TN) capture correct predictions of no increase. False Positives (FP) arise when an increase is incorrectly predicted, and False Negatives (FN) indicate missed upward movements. Accuracy (ACC) measures the proportion of correct predictions for the closing price direction.
\item \textbf{Matthews Correlation Coefficient}

Matthews Correlation Coefficient(MCC) is regarded as a balanced measure for binary classification tasks, particularly useful when class imbalance may be present. The MCC is calculated as follows:
\begin{equation}
\small\text{MCC} =
\frac{\text{TP} \cdot \text{TN} - \text{FP} \cdot \text{FN}}{
  \sqrt{
    \begin{multlined}[b] 
      (\text{TP} + \text{FP})(\text{TP} + \text{FN}) \\
      \cdot (\text{TN} + \text{FP})(\text{TN} + \text{FN})
    \end{multlined}
  }
}
\end{equation}
\end{itemize}

\begin{table*}[!htbp]
\small
\centering
\caption{Overview of the Dataset}
\label{tab:dataset_overview}
\resizebox{\textwidth}{!}{%
\begin{tabular}{@{}l p{4cm} c l l@{}}
\toprule
Event Type & Data Type & Source & Frequency & Amount \\
\midrule
Central Bank Policy \& Comms & PDF/HTML & Fed, ECB, etc. & Various & 280 \\
Govt \& Intl Macro Reports & PDF/HTML & IMF, G20, etc. & Various & 61 \\
Key Macroeconomic Data Releases & PDF/HTML/Text & BLS, U.S. PPI, etc. & Various & 2128 \\
Stock daily prices (OHLCV) & CSV & Yahoo Finance, Bloomberg, etc. & Daily & 1023 \\
XAUUSD daily price (OHLCV) & CSV & Investing.com, TradingView, etc. & Daily & 1035 \\
Industry and Company News & PDF/HTML & Reuters, Bloomberg News, etc. & Daily & 21329 \\
Current Affairs & PDF/HTML & AP, Reuters, Bloomberg, etc. & Daily & 22274 \\
\bottomrule
\end{tabular}%
}
\end{table*}

\subsection{Dataset Overview}
\label{app:dataset}
This part provides a detailed description of the curated dataset used in this paper, designed to support financial reasoning and event-aware decision-making. The dataset integrates multi-dimensional information to offer a comprehensive view of the global economic landscape, including macroeconomic indicators, monetary and fiscal policy events, financial market data, industry-specific developments, and expert investment knowledge. These data sources collectively enable structured modeling of market behavior, risk perception, and decision dynamics across multiple temporal and contextual dimensions. The dataset is organized into the following categories:

\begin{itemize}
    \item \textbf{Major Central Bank Communications and Monetary Policy Events:} This category captures policy signals from major central banks worldwide. It includes:
    \begin{itemize}
        \item Meeting Outcomes (e.g., Federal Funds Rate Decision, ECB Main Refinancing Rate)
        \item Post-Meeting Statements (e.g., FOMC Statement, ECB Press Release)
        \item Meeting Minutes (e.g., FOMC Minutes, BoE MPC Minutes)
        \item Economic Projections (e.g., U.S. SEP, BoJ Outlook Report)
        \item Official Speeches and Press Conferences (e.g., Fed Chair Remarks, ECB President Speeches)
    \end{itemize}

    \item \textbf{International Organization and Government Macroeconomic Reports:} Includes global institutional meetings and national fiscal policy disclosures:
    \begin{itemize}
        \item IMF/World Bank Meetings (e.g., WEO, GFSR)
        \item G7/G20 Finance Meetings and Communiqués
        \item National Fiscal Announcements (e.g., U.S. Federal Budget, UK Autumn Statement)
    \end{itemize}

    \item \textbf{Key Macroeconomic Data Releases:} A comprehensive set of high-frequency economic indicators, including:
    \begin{itemize}
        \item Inflation: CPI, PPI, PCE (U.S., Euro Area, China)
        \item Labor Market: Unemployment Rate, Payrolls, Earnings, Job Openings
        \item Growth: GDP, Industrial Production, Retail Sales, Durable Goods Orders
        \item Sentiment: PMI, ISM, Consumer Confidence Indices
        \item Housing: Housing Starts, Permits, Home Sales
        \item Trade: Trade Balance (U.S., Euro Area, Germany)
    \end{itemize}

    \item \textbf{Stock Price Time Series (OHLCV):} Daily Open, High, Low, Close, and Volume data for selected equities (e.g., AAPL, TSLA, XOM) from 2020--2023, sourced from Yahoo Finance and Bloomberg.

    \item \textbf{Industry and Company News:} Daily reports (PDF/HTML) on earnings, executive changes, product launches, M\&A, and regulatory updates. Sourced from Bloomberg, Reuters, WSJ, FT, etc. Includes 21,329 entries (2020--2023).

    \item \textbf{Current Affairs News:} Daily global news (2020--2023) across politics, military, economy, and society, from major agencies (AP, AFP, BBC, NYT, Xinhua). Includes 22,274 entries.

    \item \textbf{Legendary Investors’ Knowledge Base:} Structured corpus of investment philosophy and strategies from prominent investors (e.g., Buffett, Dalio, Soros), including books, interviews, letters, and speeches.
\end{itemize}

\begin{itemize}
\item \textbf{Foundational Books}: Full texts authored by or about legendary investors that codify their investment philosophy, market views, and life experiences. These include:
\begin{itemize}
\item The Intelligent Investor (Benjamin Graham)
\item Security Analysis (Graham \& Dodd)
\item One Up on Wall Street, Beating the Street (Peter Lynch)
\item Poor Charlie's Almanack (Charlie Munger)
\item The Most Important Thing (Howard Marks)
\item The Alchemy of Finance, The New Paradigm for Financial Markets (George Soros)
\item Reminiscences of a Stock Operator (Edwin Lefèvre, based on Jesse Livermore)
\item Common Stocks and Uncommon Profits (Philip Fisher)
\item The Essays of Warren Buffett (compiled by Lawrence Cunningham)
\item Margin of Safety (Seth Klarman)
\item Stocks for the Long Run (Jeremy Siegel)
\item The Dao of Capital (Mark Spitznagel)
\item The Dhandho Investor (Mohnish Pabrai)
\end{itemize}

\item \textbf{Shareholder Letters and Memos}: These writings are primary-source reflections of investors’ thinking in real time, often revealing their decision-making logic under changing market conditions:
\begin{itemize}
\item Berkshire Hathaway Shareholder Letters (Warren Buffett, 1977–present)
\item Oaktree Capital Memos (Howard Marks, archived by date and theme)
\item Daily Journal Shareholder Meetings (Charlie Munger, transcript and Q\&A)
\item Pershing Square Letters (Bill Ackman)
\item Third Point Letters (Daniel Loeb)
\item Greenlight Capital Letters (David Einhorn)
\end{itemize}

\item \textbf{Interviews, Transcripts, and Oral Histories}: These capture dynamic exchanges, spontaneous reasoning, and human intuition in investment contexts:
\begin{itemize}
\item Soros on Soros: Staying Ahead of the Curve (extended Q\&A format)
\item Warren Buffett on CNBC/Bloomberg/Fortune Interviews (indexed by date)
\item Charlie Munger Interviews with Stanford, Caltech, Redlands
\item Peter Lynch on PBS, Fidelity archives
\item Howard Marks Fireside Chats and MasterClass-style interviews
\end{itemize}

\item \textbf{Biographies and Third-party Analyses}: These offer synthesized insights into the habits, failures, and competitive edges of key investors, often triangulated with data:
\begin{itemize}
\item The Snowball (Alice Schroeder, biography of Warren Buffett)
\item More Money Than God (Sebastian Mallaby, covers Soros, Jones, Griffin, etc.)
\item The Sages: Buffett, Soros, Volcker and the Maelstrom of Markets (Charles Morris)
\item When Genius Failed (Roger Lowenstein, on LTCM \& John Meriwether)
\item The Big Short and Liar’s Poker (Michael Lewis, featuring Paulson, Burry, etc.)
\end{itemize}

\item \textbf{Public Speeches and Panel Appearances}: Transcripts and recordings from conferences, universities, or shareholder meetings, often unrehearsed and insight-rich:
\begin{itemize}
\item Berkshire Hathaway Annual Meeting Q\&A Archives (1994–present)
\item Charlie Munger's “USC Commencement Speech” and “The Psychology of Human Misjudgment”
\item Peter Thiel at Stanford and YC Startup School (macro-venture angle)
\item Ray Dalio’s “Principles” presentations at TED, Davos, IMF
\end{itemize}

\item \textbf{Documentaries and Archival Footage}: Visual content offering non-verbal cues and emotional tone:
\begin{itemize}
\item Becoming Warren Buffett (HBO Documentary)
\item Inside Job (Charles Ferguson)
\item Trader (rare Paul Tudor Jones documentary)
\item The Ascent of Money (Niall Ferguson, includes Soros and others)
\end{itemize}

These materials form the empirical backbone of a structured Human Expertise Knowledge Base. They enable the modeling of investor archetypes, simulation of investment decision processes, and empirical studies of long-term behavioral consistency and adaptive learning under uncertainty.

\end{itemize}

\subsection{Experimental details}
\subsubsection{Data Split and Evaluation Periods}
\label{app:datasplit}
To ensure the robustness and effectiveness of our model in a real-world financial environment, we adopted a rigorous time-series split for our training and test sets, explicitly considering a 7-day lookback period requirement for both OHLCV and text data.Our complete dataset spans financial market information from December 26, 2019, to December 29, 2023. This date range ensures that sufficient historical lookback data is available for the earliest predictions.
\begin{itemize}
    \item \textbf{Data Usage During the Training Phase}The model's actual training and learning, specifically the OHLCV target prediction period, is from January 2, 2020, to December 29, 2022. To make predictions for each trading day within this period, the model incorporates 7 days of preceding OHLCV and text data as historical context. This means that the OHLCV data inputs and text data inputs for the training phase range from December 26, 2019, up to December 29, 2022. Specifically, for any given prediction day within the training period (e.g., January 2, 2020), the model takes as input the OHLCV and text data from the preceding 7 days (i.e., December 26, 2019, to January 1, 2020) to learn and predict the market dynamics of January 2, 2020. This mechanism ensures that all input information (whether OHLCV or text) available to the model during training is strictly historical relative to the prediction day, rigorously preventing future information leakage.
    \item \textbf{Data Usage During the Testing Phase}After training, the model's performance is evaluated on an independent and entirely unseen OHLCV period, which runs from January 2, 2023, to December 29, 2023. Consistent with the training phase, to make predictions for each trading day within this testing period, the model also accesses its preceding 7 days of OHLCV and text information. Therefore, the OHLCV data inputs and text data inputs for the testing phase range from December 26, 2022, up to December 29, 2023.
\end{itemize}

\subsubsection{Construct Human Expertise Base}
\label{app:human expertise}
To construct the original expert knowledge base, we first collect textual data associated with the aforementioned domain experts. We then design task-specific prompts to systematically abstract structured query–knowledge pairs from the unstructured corpus,as shown in ~\ref{prompt:buffett},using Warren Buffet as an example. These prompts are executed via the \textbf{GPT-4o} (version: 2024-08-06) API, enabling the extraction of high-level conceptual queries and their corresponding expert-grounded knowledge. This procedure facilitates the transformation of raw expert content into a structured and queryable format suitable for downstream reasoning tasks.
 \subsubsection{Refine Human Expertise via verification mechanism}To ensure the reliability of the expert knowledge base, we implement a verification mechanism during the testing phase. Specifically, daily news articles are processed to extract real-time conceptual queries, which are then embedded using OpenAI's \textbf{text-embedding-3-large} model. These embeddings are compared against the pre-encoded queries in the knowledge base using cosine similarity. If a sufficiently aligned query is identified—based on a predefined similarity threshold—the corresponding expert knowledge is activated within the \textbf{Human Expertise Agent} for prediction.Successful predictions serve to validate the correctness and applicability of the retrieved expert knowledge. In cases where predictions fail, an adaptive refinement process is triggered. This process leverages feedback information collected by the \textbf{Temporal Refine Agent}, which captures contextual cues and decision inconsistencies from the failed prediction. This feedback is then incorporated into a prompt and sent to GPT-4o to generate a revised or augmented version of the expert knowledge entry. Through this iterative refinement loop, the knowledge base is further enriched and aligned with the demands of real-world market conditions, thereby enhancing the original human expert knowledge.
\subsubsection{Detailed Configurations in Experiments}
\begin{itemize}
    \item \textbf{Trend Forecasting Task Configurations}
    \label{app:forecasting config}
    To ensure a fair comparison for the Trend Forecasting task, all models, including FinHEAR and the various baselines, were configured to output a binary prediction indicating the next-day directional movement of the asset (e.g., "Up" or "Down").

    For classification-based baselines, including ALSTM, Informer, and StockNet, their final output layers were designed for two-class prediction. This layer consists of a fully connected layer with two output units followed by a softmax activation, with each class corresponding to either an "Up" or "Down" price movement. At each time step, the model outputs a probability distribution over these two potential outcomes, and the ultimate trend forecast is derived by selecting the class with the highest probability. This standardized two-class output setup ensures consistency in the prediction space across these models.
    
    Similarly, for Reinforcement Learning (RL) based methods such as PPO and DQN, their action spaces were adapted to a binary classification for this task. Instead of continuous position sizing or detailed trading actions, these agents learn to output one of two discrete actions: predicting "Up" or predicting "Down." The reward function in this context is tailored to maximize the accuracy of these directional predictions.
    
    For Large Language Model (LLM) based methods, including the COT agent, GA generative agent, FinMem, and FinCon, their output generation was constrained to produce a binary forecast. At each decision time step, these agents were prompted or fine-tuned to extract a clear "Up" or "Down" prediction from their text output using a parsing module. This ensures that their output directly corresponds to the two directional classes required for this task.
    
    Through this consistent binary output configuration across all models, we can perform a comprehensive and equitable performance comparison specifically for the trend forecasting task.
    \item \textbf{Trading Task Configurations}
    \label{app:trading config}
    To ensure a fair comparison with the FinHEAR model, which is capable of selecting from four distinct trading actions at each decision point—Long, Short, Hold, and Close—we configured the baseline models used in our study accordingly. For the classification-based baselines, including ALSTM, Informer, and StockNet, their final output layers were designed for four-class prediction. This layer consists of a fully connected layer with four output units followed by a softmax activation, with each class corresponding to one of the aforementioned trading actions: 'Long', 'Short', 'Hold', and 'Close'. At each trading time step, the model outputs a probability distribution over these four potential actions, and the ultimate trading signal is typically derived by selecting the action with the highest probability or score. This standardized four-class output setup guarantees consistency in the discrete action space across these models.

    Differing in their action space design, we also employed Reinforcement Learning (RL) based methods as baselines capable of more flexible position management. Specifically, we configured agents based on Proximal Policy Optimization (PPO) and Deep Q-Network (DQN). These RL agents learn a trading policy aiming to maximize a cumulative reward signal. The action space configuration varies: PPO utilizes a continuous action space, outputting a target position size within [-1, 1]; whereas DQN operates in a discrete action space, with actions encompassing key trading decisions like 'Open Long', 'Open Short', 'Close Position', and 'Hold Position'. The state space for both RL agents comprises relevant market observations (e.g., historical price series, technical indicators) and information about the agent's current portfolio status (e.g., current position, cash). The reward function guiding the learning process is based on portfolio performance metrics, such as the change in portfolio value or risk-adjusted return over each time step.

    As another category of baselines capable of more flexible position management, we included Large Language Model (LLM) based methods, such as the COT agent, GA generative agent, FinMem, and FinCon. These methods leverage LLMs to analyze market data and generate trading recommendations for continuous position management. At each decision time step, these agents accurately extract a concrete, continuous target position size from the text output generated by the LLM using a parsing module. This continuous target position size directly dictates the agent's trading behavior and intent: a positive target position represents a Long intent; a negative target position indicates a Short intent; and a zero target position corresponds to a Close intent (when holding a position) or maintaining a cash position (Hold) intent (when not holding a position).

    Through these diverse baseline configurations spanning different action space granularities, we are able to perform a comprehensive and equitable performance comparison relative to FinHEAR.
\end{itemize}

\subsubsection{Single Asset Trading Result Graphs}
\label{app:singleasset}
\begin{figure*}[]
    \centering
    \begin{subfigure}[b]{0.45\textwidth}
        \centering
        \includegraphics[width=\linewidth]{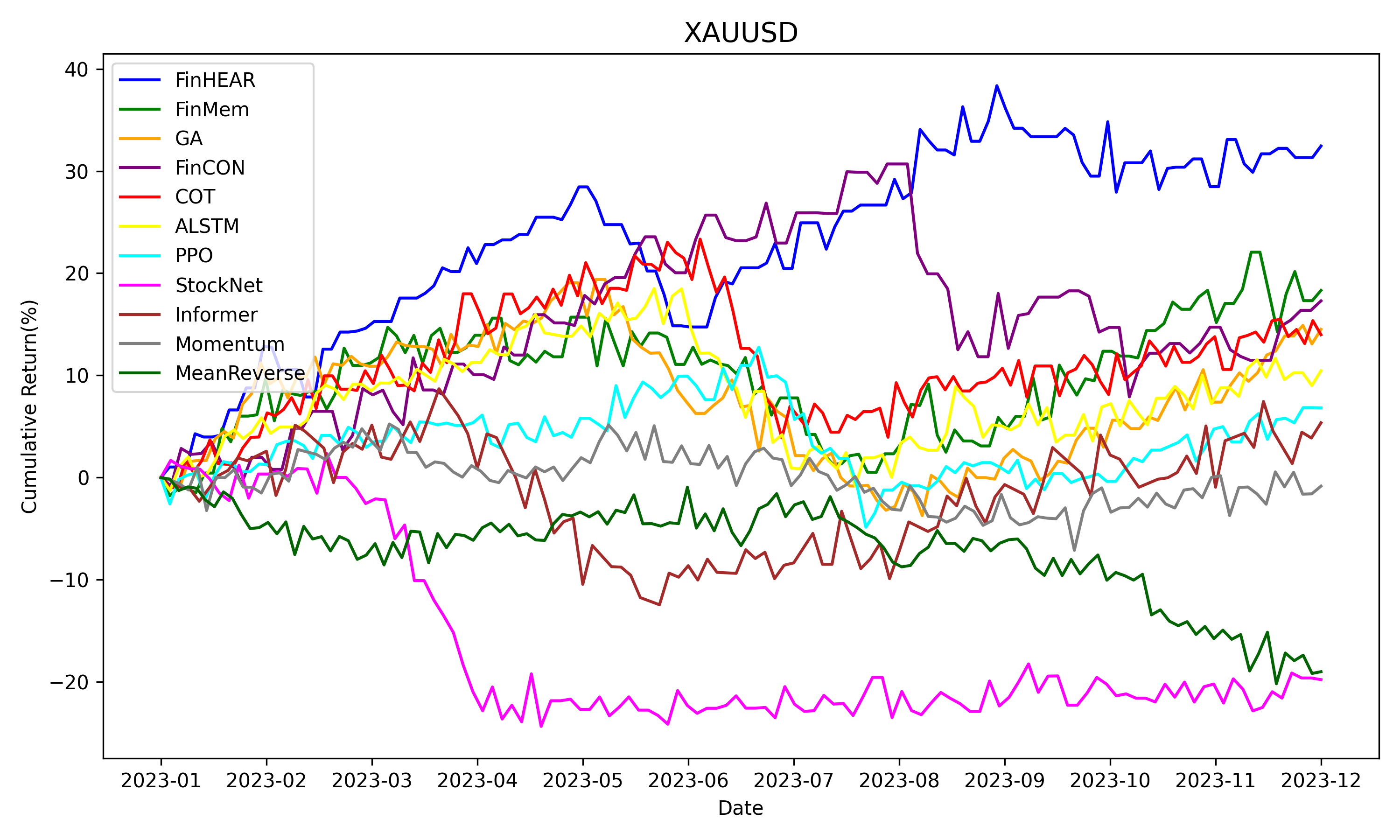}
        \caption{XAUUSD}
        \label{fig:xauusd}
    \end{subfigure}
    \hfill
    \begin{subfigure}[b]{0.45\textwidth}
        \centering
        \includegraphics[width=\linewidth]{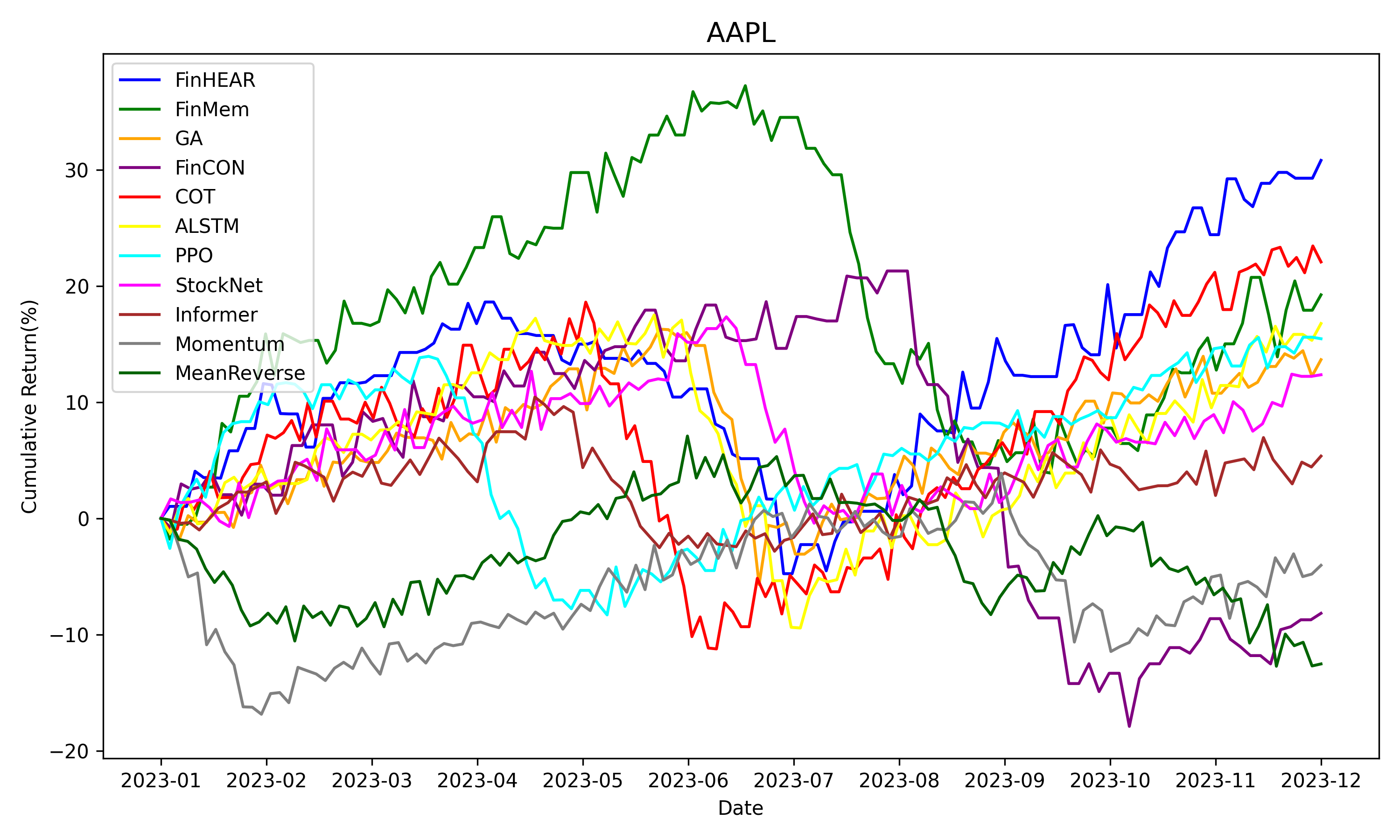}
        \caption{AAPL}
        \label{fig:aapl}
    \end{subfigure}

    \vspace{0.5em}

    \begin{subfigure}[b]{0.45\textwidth}
        \centering
        \includegraphics[width=\linewidth]{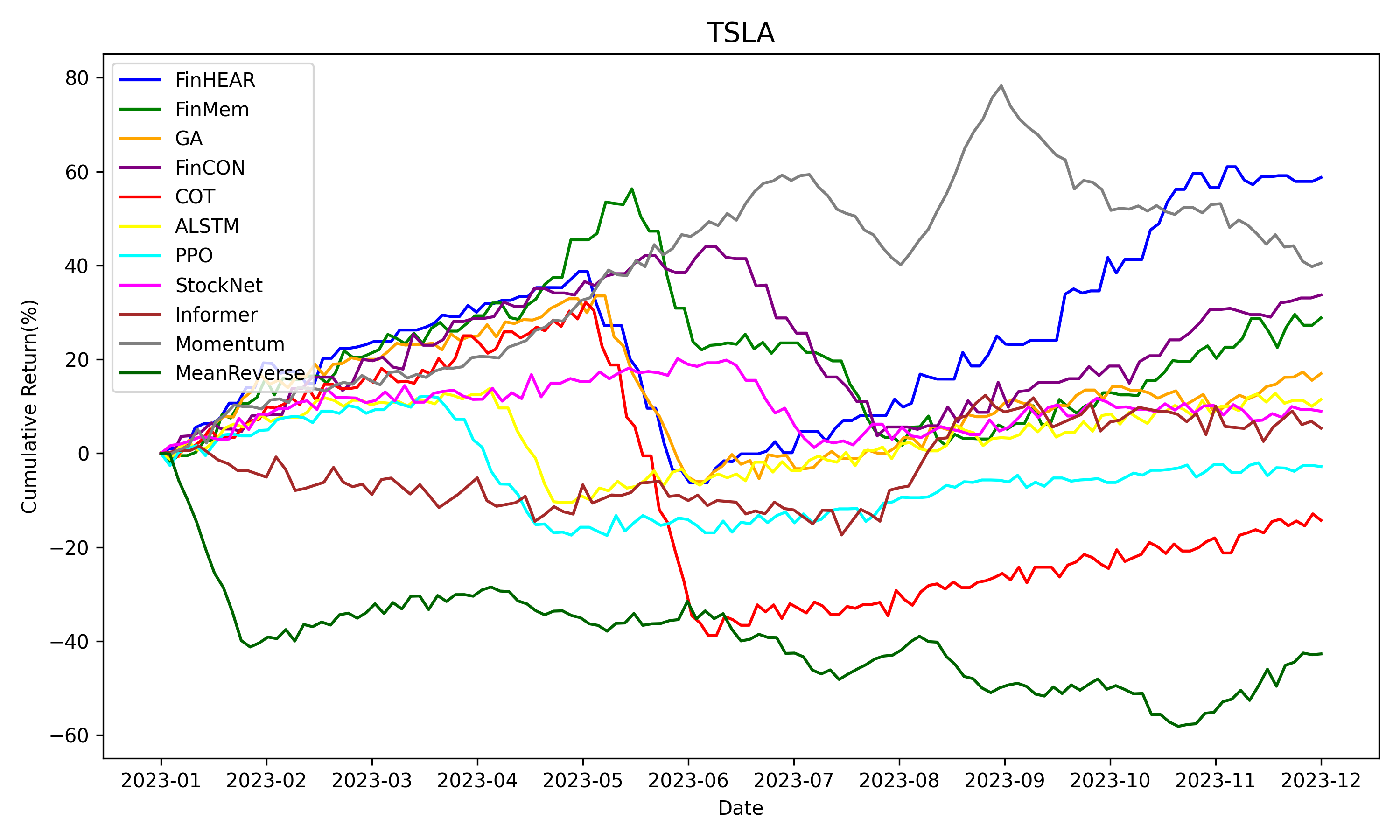}
        \caption{TSLA}
        \label{fig:tsla}
    \end{subfigure}
    \hfill
    \begin{subfigure}[b]{0.45\textwidth}
        \centering
        \includegraphics[width=\linewidth]{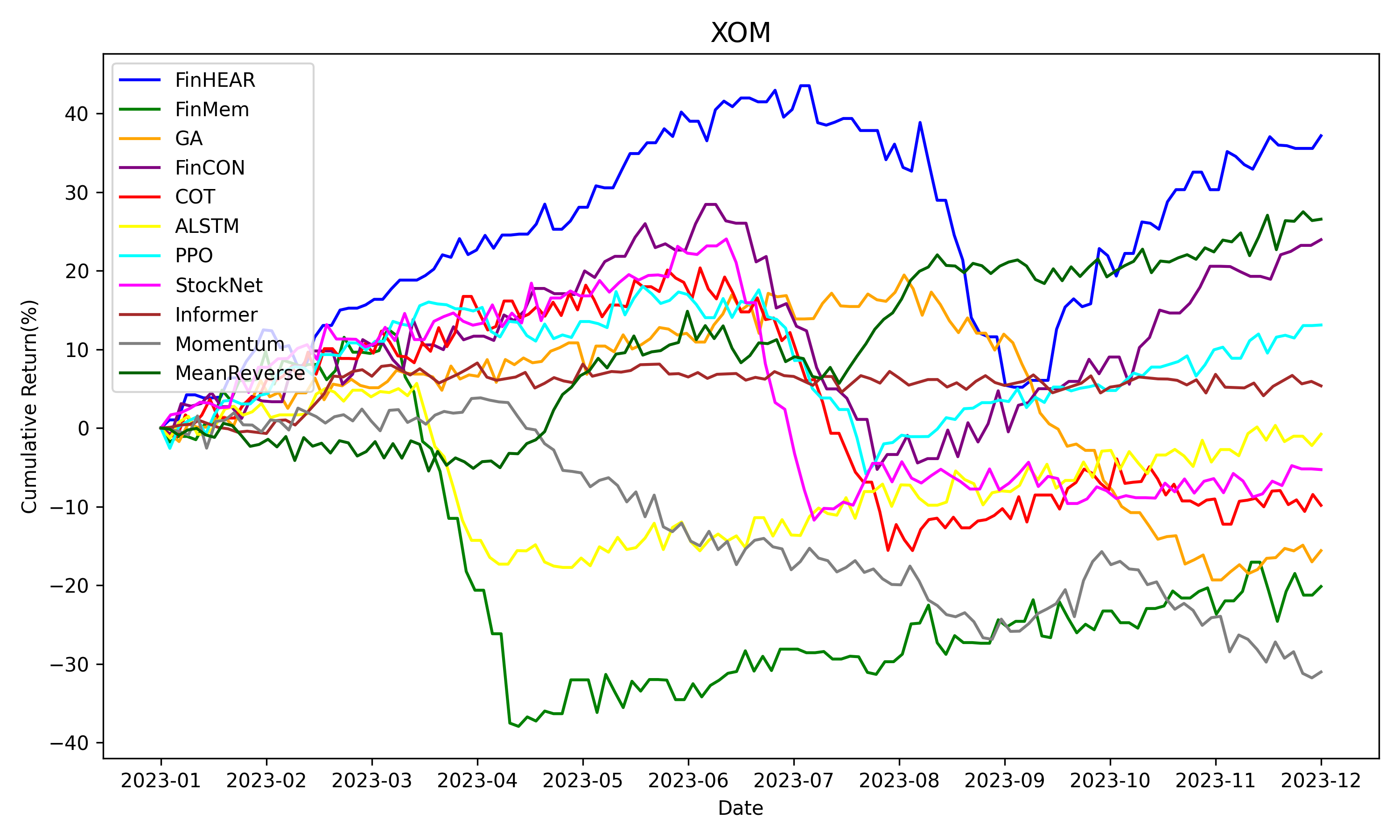}
        \caption{XOM}
        \label{fig:xom}
    \end{subfigure}

    \caption{Cumulative trading performance of FinHEAR and other baselines on four individual assets (XAUUSD, AAPL, TSLA, XOM) from January 2023 to December 2023.}
    \label{fig:single_asset_performance}
\end{figure*}

as shown in Table~\ref{fig:xauusd},Table~\ref{fig:aapl},Table~\ref{fig:tsla},Table~\ref{fig:xom},we visualize the cumulative return trajectories for four individual assets—XAUUSD, AAPL, TSLA, and XOM—throughout 2023. These line charts reflect the trading performance using cumulative return as the primary evaluation metric.

\begin{figure*}[]
    \centering
    \begin{subfigure}[b]{0.45\textwidth}
        \centering
        \includegraphics[width=\linewidth]{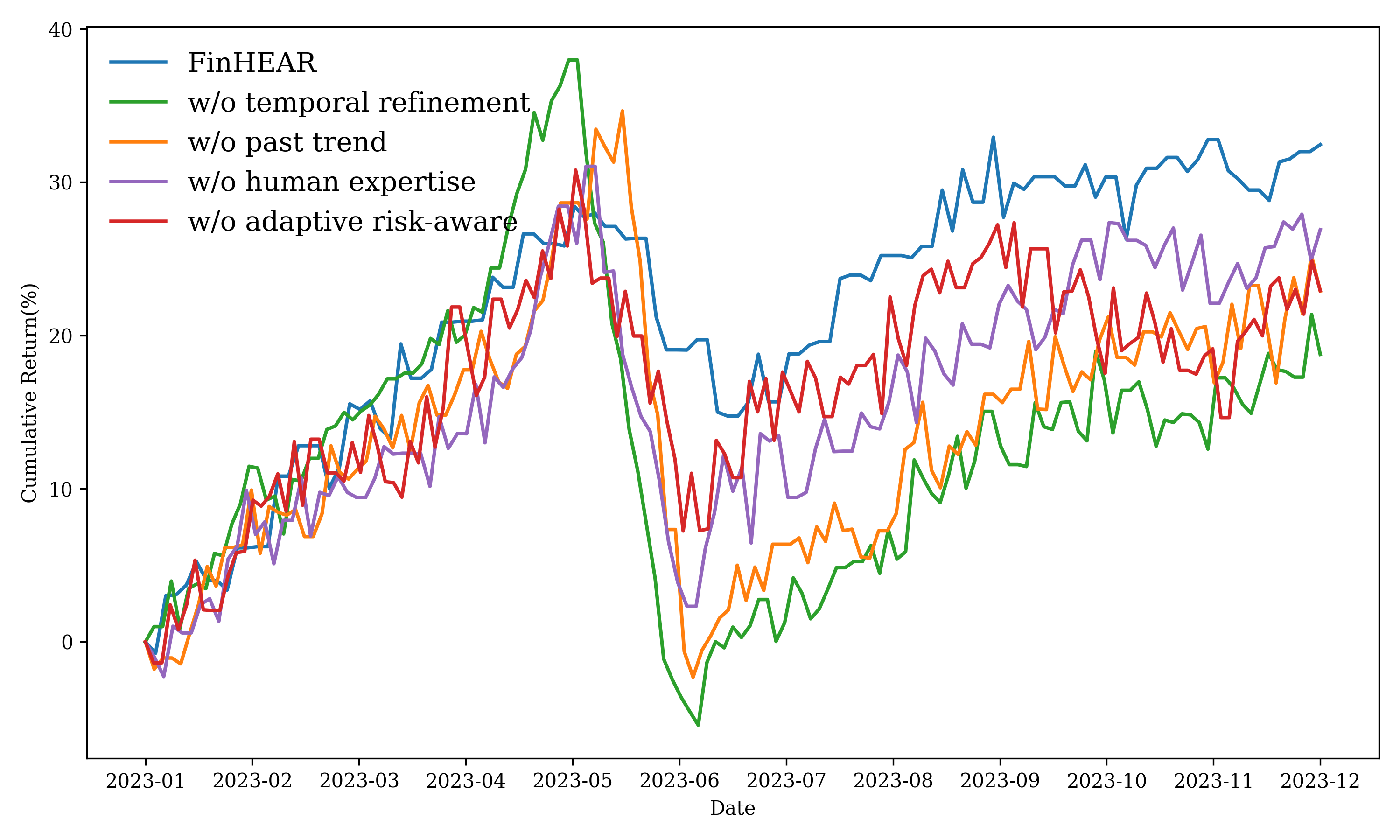}
        \caption{XAUUSD}
        \label{fig:abl_xauusd}
    \end{subfigure}
    \hfill
    \begin{subfigure}[b]{0.45\textwidth}
        \centering
        \includegraphics[width=\linewidth]{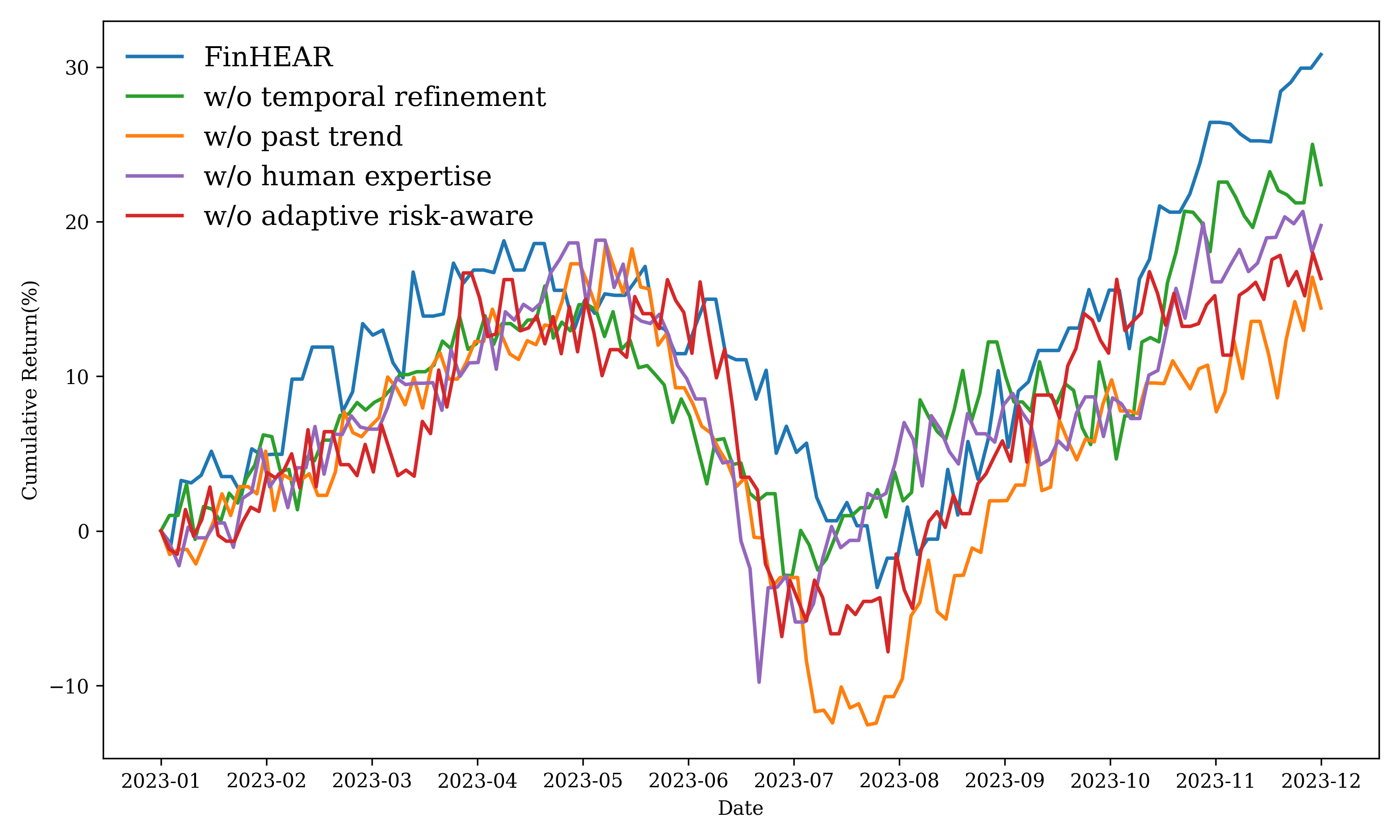}
        \caption{AAPL}
        \label{fig:abl_aapl}
    \end{subfigure}

    \vspace{0.5em}

    \begin{subfigure}[b]{0.45\textwidth}
        \centering
        \includegraphics[width=\linewidth]{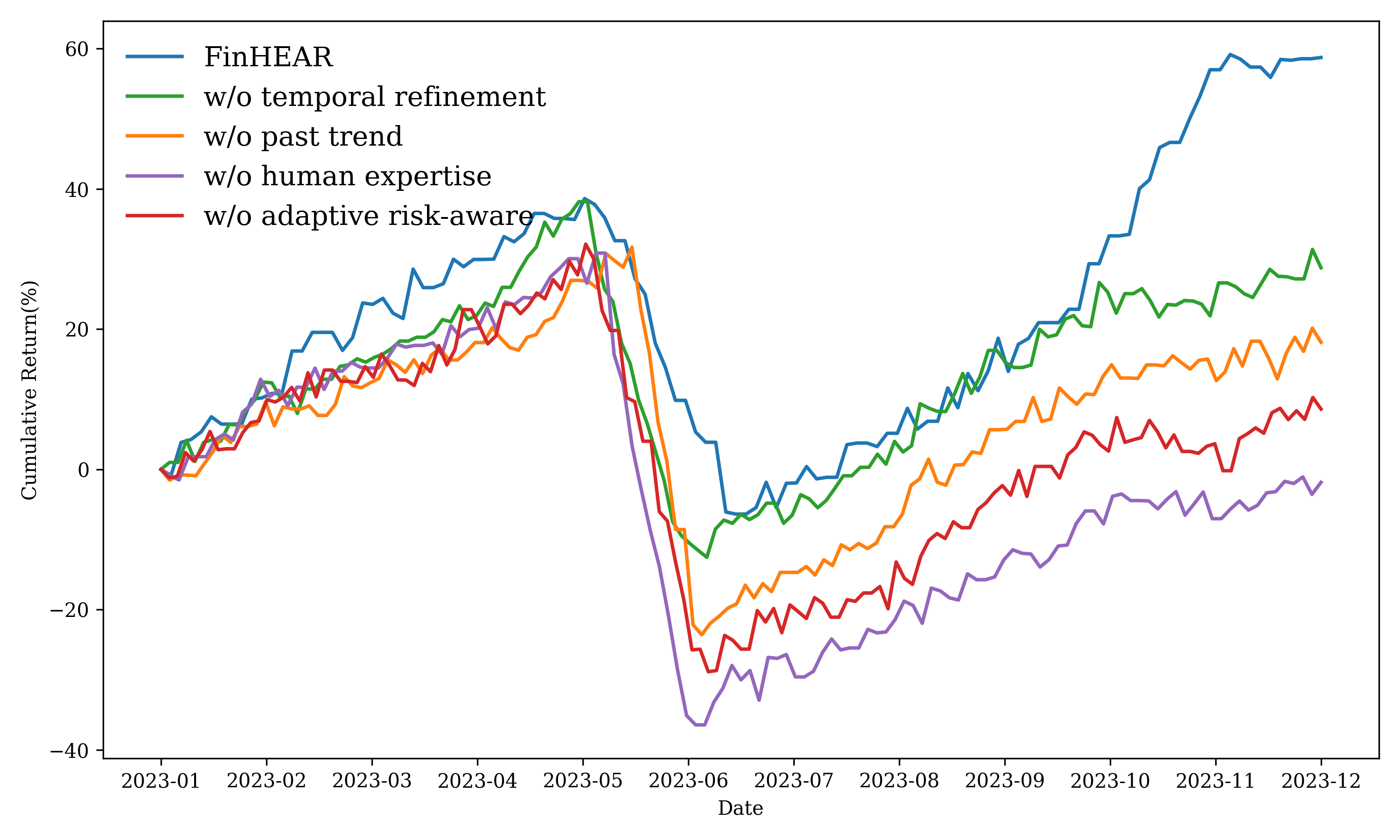}
        \caption{TSLA}
        \label{fig:abl_tsla}
    \end{subfigure}
    \hfill
    \begin{subfigure}[b]{0.45\textwidth}
        \centering
        \includegraphics[width=\linewidth]{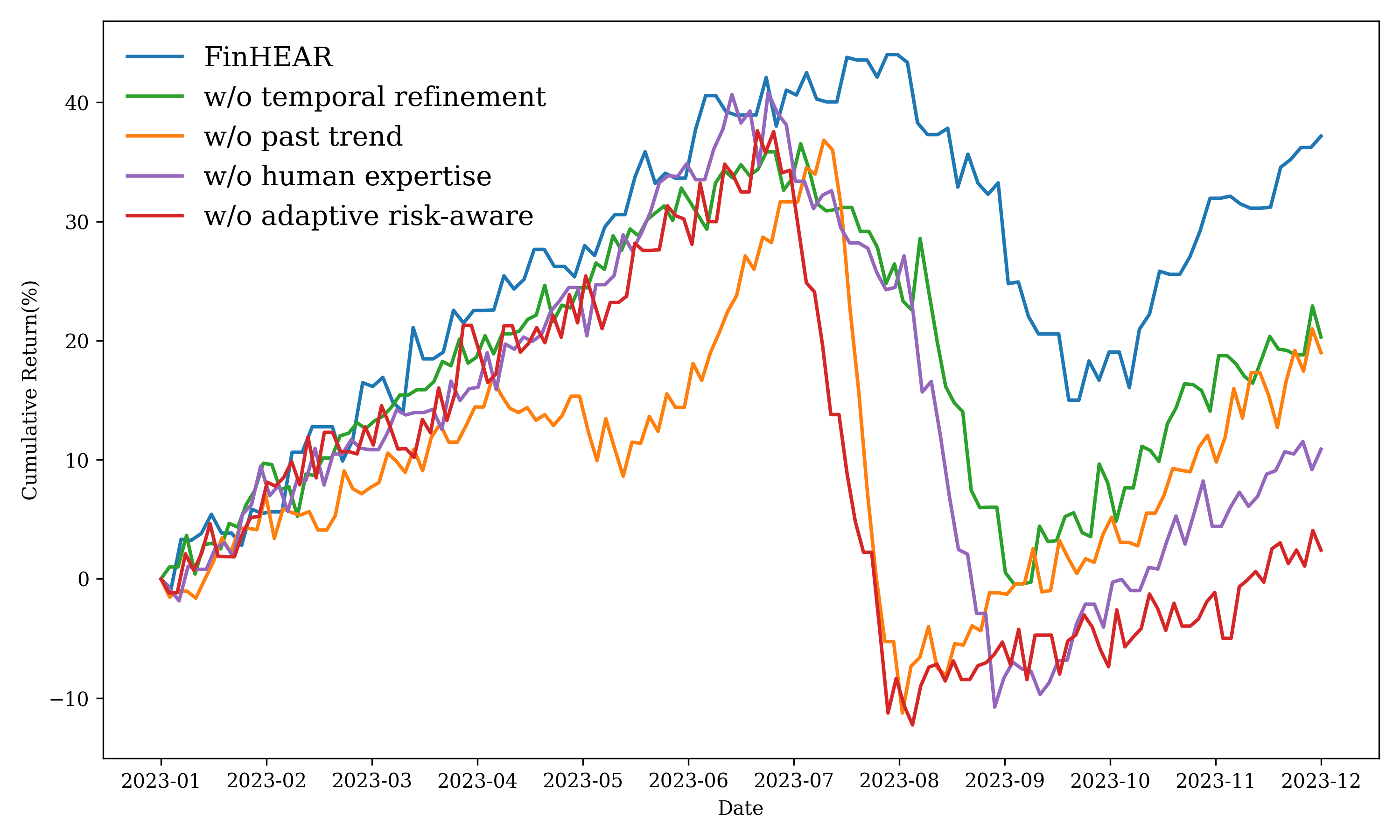}
        \caption{XOM}
        \label{fig:abl_xom}
    \end{subfigure}

    \caption{Cumulative trading performance of FinHEAR and its ablation variants on  four individual assets (XAUUSD,AAPL, TSLA, XOM) from January 2023 to December 2023}
    \label{fig:single_asset_ablation_performance}
\end{figure*}

\subsubsection{Impact of Transaction Costs}
\label{app:transaction_costs}

To ensure our results are not an artifact of ignoring trading frictions, we conducted a comprehensive backtest under conservative transaction cost assumptions:

\begin{enumerate}
    \item \textbf{New Backtest with Transaction Costs}  
    Following reviewer suggestions, we repeated the AAPL trading experiments, incorporating realistic transaction costs across all methods for a fair comparison.
    
    \item \textbf{Conservative Cost Assumption}  
    For every transaction (both entries and exits), we applied a transaction cost of \textbf{0.05\%} (5 basis points) per side, resulting in a total round-trip cost of \textbf{0.1\%}. This rate accounts for typical brokerage commissions and average slippage.
    
    \item \textbf{Results on AAPL: Performance Advantage Remains Robust}  
    As shown in Table~\ref{tab:transaction_costs}, while transaction costs reduce absolute returns across all strategies, \textsc{FinHEAR} continues to significantly outperform all strong baselines in both Cumulative Return (\textbf{CR}) and Sharpe Ratio (\textbf{SR}). This demonstrates that our framework's ability to generate alpha is not dependent on unrealistic cost assumptions.
\end{enumerate}

\begin{table}[h]
\centering
\small
\setlength{\tabcolsep}{5pt} 
\renewcommand{\arraystretch}{1.15} 
\begin{tabular}{lcccc}
\toprule
\multirow{2}{*}{\textbf{Method}} & \multicolumn{2}{c}{\textbf{No Costs}} & \multicolumn{2}{c}{\textbf{With Costs}} \\
\cmidrule(lr){2-3} \cmidrule(lr){4-5}
 & \textbf{CR} & \textbf{SR} & \textbf{CR} & \textbf{SR} \\
\midrule
\textsc{FinHEAR} & \textbf{30.81} & \textbf{1.95} & \textbf{25.96} & \textbf{1.68} \\
ALSTM            & 16.77 & 0.75 & 8.97 & 0.41 \\
Informer         & 8.54  & 1.22 & 5.06 & 0.82 \\
PPO              & 15.46 & 1.67 & 10.33 & 1.01 \\
FinCon           & -8.17 & -0.53 & -13.32 & -1.62 \\
FinMem           & 19.24 & 1.43 & 15.09 & 1.19 \\
CoT              & 22.07 & 1.32 & 16.14 & 1.28 \\
\bottomrule
\end{tabular}
\caption{Impact of transaction costs on trading performance for the AAPL task. CR = Cumulative Return, SR = Sharpe Ratio. \textsc{FinHEAR} retains a significant advantage even under realistic cost assumptions.}
\label{tab:transaction_costs}
\end{table}

\subsubsection{Ablation}
\label{app:ablation}
as shown in Table~\ref{tab:asset_performance},based on the analysis of asset performance across the different time windows presented in the table, we observe a consistent pattern: various financial asset classes tend to achieve their peak relative performance over distinct time horizons. This differentiation is not coincidental but is intrinsically linked to their inherent asset characteristics and the temporal scale at which their core value drivers exert influence.

Assets whose value is highly contingent upon short-term market sentiment, immediate events, or rapid catalysts—such as gold (XAUUSD), which reacts swiftly to short-term uncertainty as a safe haven, or high-growth, high-volatility equities like Tesla (TSLA), prone to significant surges driven by recent favorable news or speculative momentum—tend to exhibit their most robust performance concentrated within shorter time windows. The performance of these assets is often characterized by high volatility in the short term, and sustaining such peak performance when averaged over extended time horizons can be challenging.

In contrast, optimal performance for some assets may manifest within a medium-term window, as exemplified by Apple (AAPL). The value drivers for this category of assets are typically more closely tied to factors such as product life cycles, consistent earnings growth, and medium-term industry trends. The impact of these factors requires a certain duration (typically several quarters to a few years) to fully materialize and be reflected in asset valuation. While extremely short windows may not adequately capture these cumulative effects, overly long horizons might introduce challenges such as growth deceleration or shifts in market cycles.

Furthermore, the performance of other assets is more likely to reach its optimum or show continuous improvement over longer time windows, with ExxonMobil (XOM) serving as an illustrative example. This class of assets is often closely linked to the long-term cycles of the macroeconomy, persistent trends in major commodity prices, or structural shifts in supply and demand. The evolution of these influencing factors is typically gradual and sustained, requiring several years to fully unfold. Over longer cycles, the impact of short-term market noise becomes diluted, and the asset's underlying cyclicality or fundamental long-term trend becomes dominant, enabling it to achieve superior cumulative performance.

From a financial theory perspective, this phenomenon highlights the concept of 'Temporal Specificity of Asset Performance' – where the dominant time scale of an asset's core value drivers aligns effectively with the investment or observation horizon. Understanding this alignment is crucial for investors seeking to grasp the risk-return characteristics of various assets and to consider the temporal matching when constructing investment portfolios.
\begin{table}[htbp]
\centering
\caption{Asset performance (\%) over different time windows}
\label{tab:asset_performance}
\small
\begin{tabular}{lccc}
\toprule
\textbf{Asset} & \textbf{Window = 3} & \textbf{Window = 5} & \textbf{Window = 7} \\
\midrule
XAUUSD & \cellcolor{gray!10}\textbf{48.93} & 32.45 & 22.46 \\
AAPL   & 29.15 & \cellcolor{gray!10}\textbf{30.81} & 23.56 \\
TSLA   & \cellcolor{gray!10}\textbf{63.82} & 58.47 & 40.13 \\
XOM    & 35.19 & 37.16 & \cellcolor{gray!10}\textbf{45.53} \\
\bottomrule
\end{tabular}
\end{table}

as shown in Figure ~\ref{fig:abl_xauusd},Figure ~\ref{fig:abl_aapl},Figure ~\ref{fig:abl_tsla}and Figure ~\ref{fig:abl_xom}, we present the cumulative trading performance of FinHEAR and its ablation variants on four individual assets: XAUUSD, AAPL, TSLA, and XOM. These ablation studies are conducted to evaluate the contribution of different key components of the FinHEAR model to its overall trading performance in single-asset markets. The figures illustrate the cumulative returns achieved by the full FinHEAR model compared to its variants with specific modules removed over the period from January 2023 to December 2023 for each respective asset, highlighting the effectiveness of each component.

\subsubsection{Agent Modular Design and Prompt Engineering}    
This appendix provides a detailed exposition of representative prompt templates employed for the specialized agents within FinHEAR, serving to illustrate our approach to encoding distinct functionalities and expert reasoning paradigms. Each template is meticulously crafted to guide the Large Language Models (LLMs) in performing specific analytical and reasoning tasks, ensuring both fidelity to the intended behavioral profiles and structured output generation.

\begin{figure*}[hb]

\centering
\begin{tcolorbox}[
    width=0.95\textwidth,
    colframe=black!60,
    colback=gray!5,
    fonttitle=\bfseries,
    fontupper=\footnotesize,
    title={Prompt Template: Buffett-style Investment Reasoning},
    breakable,
    enhanced,
    sharp corners=southwest,
    boxrule=0.5pt,
    left=2mm, right=2mm, top=1mm, bottom=1mm
]

\textbf{Input:}  

The input is a segment of natural language text, denoted as \texttt{\{text\}}.  
This passage is assumed to discuss \textbf{Warren Buffett}, one of the most influential value investors of the 20th and 21st centuries.  
Buffett is widely known for his adherence to fundamental investing principles including, but not limited to: intrinsic value assessment, long-term business fundamentals, margin of safety, rational capital allocation, and resistance to market sentiment.

\textbf{Task Specification:}  

Your objective is two-fold:  
\begin{itemize}
    \item Abstract the core content of the text into a reusable, generalized investment reasoning query or problem formulation (denoted as \texttt{Query}). This abstraction should reflect a broader class of investment reasoning problems that the input exemplifies.  
    \item Distill a reasoning trajectory that emulates Warren Buffett's investment paradigm. This reasoning process should reflect deliberate, logically grounded, and value-aligned investment thinking, referencing concepts such as economic moats, discounted cash flow analysis, long-term competitive advantage, and probabilistic evaluation of downside risk.
\end{itemize}

\textbf{Output Constraints:}
\begin{itemize}
    \item You must produce one or more outputs strictly following the syntactic template specified below.
    \item Each output unit must maintain the structure exactly, and must not include any preambles, justifications, explanations, or commentary.
\end{itemize}

\textbf{Output Format:}
\begin{itemize}
    \item Query: <A generalized investment reasoning problem abstracted from the input>
    \item Human Expertise: <A detailed, logically sequenced reasoning trajectory consistent with Buffett's value investing methodology>
\end{itemize}

\textbf{Note:}
\begin{itemize}
    \item If multiple independent abstractions and reasoning chains can be derived from the input, return them as separate entries, each adhering to the exact format.
    \item The reasoning chain should demonstrate internal logical consistency and align with Buffett's documented investment philosophy (e.g., as expressed in shareholder letters, interviews, or books).
    \item Avoid surface-level or tautological restatements; prioritize deep reasoning that could generalize across investment contexts.
\end{itemize}

\end{tcolorbox}

\caption{Prompt template for extracting generalized investment reasoning in the style of Warren Buffett.}
\label{prompt:buffett}
\end{figure*}

\begin{figure*}[hb]
\label{prompt:historical}
\centering
\begin{tcolorbox}[
    width=0.95\textwidth,
    colframe=black!60,
    colback=gray!5,
    fonttitle=\bfseries,
    fontupper=\footnotesize,
    title={Historical Trend Agent Prompt Template},
    breakable,
    enhanced,
    sharp corners=southwest,
    boxrule=0.5pt,
    left=2mm, right=2mm, top=1mm, bottom=1mm
]
[Past Gate - Historical Pattern Extraction Module]

\textbf{Context}:
You are a financial analysis agent responsible for mining historical trading data and annotated return outcomes to extract statistically and economically significant patterns. The rate of return is defined as:
(tomorrow's closing price - today's closing price) / today's closing price.

\textbf{Input}:
\begin{itemize}
    \item Historical investment experiences with associated return labels:
    \texttt{\{past\_exps\}}
    
    \item Raw historical market data containing the following structured fields:

    - Date

    - Open

    - High

    - Low

    - Close

    - Price (if different from Close, assume it's a derived value)

    - Volume

    \texttt{\{data\}}
\end{itemize}

\textbf{Instructions}
 \begin{itemize}
     \item  Identify meaningful temporal trends, recurring signals, or anomalous movements that historically preceded above- or below-average returns.
    \item  Cross-reference return outcomes with relevant price and volume dynamics, candlestick behaviors, or volatility shifts to uncover predictive patterns.
    \item  Extract insights that may have forward-looking value for real-time investment decision-making.
    \item  Prioritize signal relevance, statistical consistency, and interpretability in your summary.
    \item Keep the tone formal, objective, and consistent with academic or institutional research standards.
 \end{itemize}
 
\textbf{Output Format:}

 Present your analytical findings strictly using the following structure. Do \textbf{not} include additional commentary or formatting beyond what is shown:
  
 [Past\_summary: <Concise yet information-rich synthesis of 2–4 sentences, outlining key historical trends, signal behaviors, or statistical regularities that may inform current trading strategy.>]

\end{tcolorbox}

\caption{Prompt Template for Current Event Agent Financial Analysis}
\end{figure*}

\begin{figure*}[hb]
\label{prompt:verfiy}
\centering
\begin{tcolorbox}[
    width=0.95\textwidth,
    colframe=black!60,
    colback=gray!5,
    fonttitle=\bfseries,
    fontupper=\footnotesize,
    title={Temporal Refinement Agent Prompt Template},
    breakable,
    enhanced,
    sharp corners=southwest,
    boxrule=0.5pt,
    left=2mm, right=2mm, top=1mm, bottom=1mm
]
[Temporal Refinement Gate]

You are tasked with validating and refining a financial market analysis for \texttt{\{date\}}.

The expected return (R) is defined as:

R = (P\_t+1 - P\_t) / P\_t

Where:
\begin{itemize}
    \item P\_t denotes today's closing price.
    \item P\_t+1 denotes tomorrow's closing price.
\end{itemize}

Below is the original market analysis and the model-generated forecast for today:\texttt{\{today\_exp\}}\texttt{\{temp\}}

Your objective is to revise the analysis, ensuring that it aligns with the actual return direction implied by the above definition.

\textbf{Guidelines:}
\begin{itemize}
    \item Revise only the analytical interpretation. Do \textbf{not} alter factual observations unless necessary for consistency.
    \item Be precise and concise.
    \item Use terminology appropriate for financial and economic analysis.
    \item Do \textbf{not} include any additional commentary, justification, or explanation.
\end{itemize}
Output must \textbf{strictly} follow the structure below (no deviations):

[\texttt{\{date\}}\_summary: <revised analysis text here>]
\end{tcolorbox}
\caption{Prompt Template for Temporal Refinement Agent Financial Analysis and Decision}
\end{figure*}

\begin{figure*}[hb]
\label{prompt:current}
\centering
\begin{tcolorbox}[
    width=0.95\textwidth,
    colframe=black!60,
    colback=gray!5,
    fonttitle=\bfseries,
    fontupper=\footnotesize,
    title={Current Event Agent Prompt Template},
    breakable,
    enhanced,
    sharp corners=southwest,
    boxrule=0.5pt,
    left=2mm, right=2mm, top=1mm, bottom=1mm
]
[Current Gate - Integrated Market Reasoning Module]

\textbf{Context:}
You are a decision-making financial agent that synthesizes real-time macroeconomic and geopolitical events with historical market patterns to generate immediate and actionable market signals. Your reasoning must be both forward-looking and grounded in empirical context.
\textbf{Input:}
\begin{itemize}
    \item Real-time event feed containing the most recent developments:
\texttt{\{current\}}
    \item Extracted historical insights and trend summaries derived from past performance and return data:
\texttt{\{past\_info\}}
\end{itemize}
\textbf{Instructions:}
\begin{itemize}
    \item Integrate current events with historical context to assess potential impacts across equity, commodity, fixed income, and currency markets.
    \item Identify dominant economic narratives (e.g., inflation risk, monetary tightening, supply chain disruptions) and connect them with previously observed market reactions.
    \item Produce a concise and high-signal analysis suitable for immediate investment positioning or portfolio adjustment.
    \item Focus only on the most actionable insights with measurable short-term implications.
    \item  Ensure analytical clarity, avoiding speculation or generalizations not supported by the data.
\end{itemize}
\textbf{Output Format:}

Your analysis must follow \textbf{exactly} the format below,\textbf{without} any introductory or concluding remarks:

[Current\_summary: <A compact, structured evaluation (~2–4 sentences) outlining the synthesis of present events with historical insights, leading to clear, justifiable market signals.>]

\end{tcolorbox}

\caption{Prompt Template for Current Event Agent Financial Analysis}

\end{figure*}

\begin{figure*}[hb]
\label{prompt:persona}
\centering
\begin{tcolorbox}[
    width=0.95\textwidth,
    colframe=black!60,
    colback=gray!5,
    fonttitle=\bfseries,
    fontupper=\footnotesize,
    title={Human Expertise Agent Prompt Template},
    breakable,
    enhanced,
    sharp corners=southwest,
    boxrule=0.5pt,
    left=2mm, right=2mm, top=1mm, bottom=1mm
]

[Persona Gate – Behavioral Adjustment \& Sentiment Integration Module]

\textbf{Context:}

You are a behavioral-aware financial reasoning agent. Your role is to adjust and refine market analyses by incorporating investor sentiment, psychological biases, and historically consistent behavior patterns observed across different market regimes.

\textbf{Input:}
\begin{itemize}
    \item Investor persona insights, including sentiment indicators and behavioral response patterns:\texttt{\{persona\}}
    \item Preliminary market analysis based on current macro-events and historical data context:
\texttt{\{current\_info\}}
\end{itemize}
\textbf{Instructions:}
\begin{itemize}
    \item Interpret how different types of investors (e.g., risk-averse, momentum-driven, contrarian) are likely to perceive and react to the current market situation.
    \item Adjust the current analysis accordingly, integrating behavioral finance theory (e.g., loss aversion, herd behavior, overreaction) to reflect more realistic market responses.

    \item Emphasize actionable implications that arise when aligning analytical signals with investor psychology.
    \item Maintain analytical rigor, and ensure output remains concise, data-grounded, and investment-oriented.
\end{itemize}
\textbf{Output Format:}

Please provide your refined analysis using the structure below, \textbf{without} adding commentary or deviation from the format:

[refined\_summary: <A revised analysis (~2–4 sentences) that merges objective market signals with plausible investor reactions, enhancing practical applicability.>]

\end{tcolorbox}

\caption{Prompt Template for Human Expertise Agent Financial Analysis}
\end{figure*}

\begin{figure*}[ht]
\label{prompt:risk}
\centering
\begin{tcolorbox}[
    width=0.95\textwidth,
    colframe=black!60,
    colback=gray!5,
    fonttitle=\bfseries,
    fontupper=\footnotesize,
    title={Risk Analysis Agent Prompt Template},
    breakable,
    enhanced,
    sharp corners=southwest,
    boxrule=0.5pt,
    left=2mm, right=2mm, top=1mm, bottom=1mm
]
[Risk Gate – Multi-Layer Analytical Consistency \& Decision Alignment Module]

\textbf{Context:}

You are a risk assessment agent responsible for evaluating the coherence, reliability, and internal alignment of a multi-stage investment analysis pipeline. Your goal is to assess the overall risk exposure associated with the final investment decision based on preceding analytical layers.

\textbf{input:}
\begin{itemize}
    \item  Historical Analysis Output (Trend-Based Retrospective Insight): \texttt{\{past\_info\}}
    \item  Current Event-Driven Market Analysis: \texttt{\{current\_info\}}
    \item  Behaviorally Adjusted (Persona-Based) Refined Analysis: \texttt{\{refined\_info\}}
    \item  Final Proposed Investment Action: \texttt{\{decison\}}
\end{itemize}
\textbf{Instructions:}

\begin{itemize}
    \item Assess the \textbf{internal consistency} across the three layers of analysis (historical, current, refined), identifying areas of alignment or contradiction in sentiment, directional bias (e.g., bullish/bearish), and strategic tone.

    \item Identify any \textbf{behavioral or analytical divergence} that could introduce decision risk (e.g., if refined analysis is bullish while past/current analyses are neutral or negative).

    \item Evaluate how well the final investment action (e.g., Buy/Sell) aligns with the overall analytical trajectory.

    \item Determine the \textbf{overall risk level} of the decision using the following classification:
    \begin{itemize}
        \item \textbf{Low}: High consistency, strong alignment, low analytical or behavioral divergence.
        
        \item \textbf{Medium}: Moderate consistency, partial misalignment, or behavioral ambiguity.
        
        \item \textbf{High}: Conflicting signals, sentiment contradiction, or misaligned final action.

    \end{itemize}
\end{itemize}

\textbf{Output Format:}

Respond strictly using the following structure, with no commentary outside the format:

[risk\_level: <Low or Medium or High>, risk\_evaluation: <Concise and reasoned explanation (~2–4 sentences) summarizing alignment, conflict, and overall risk considerations.>]

\end{tcolorbox}

\caption{Prompt Template for Rise Analysis Agent Financial Analysis}
\end{figure*}

\begin{figure*}[hb]
\label{prompt:alignment}
\centering
\begin{tcolorbox}[
    width=0.95\textwidth,
    colframe=black!60,
    colback=gray!5,
    fonttitle=\bfseries,
    fontupper=\footnotesize,
    title={Expert Alignment Agent Prompt Template},
    breakable,
    enhanced,
    sharp corners=southwest,
    boxrule=0.5pt,
    left=2mm, right=2mm, top=1mm, bottom=1mm
]
[Expert Alignment Gate – Multi-Agent Decision Validation Layer]

\textbf{Role:}

You are a centralized decision-making controller operating within a multi-agent investment advisory system. Each expert agent is specialized in a particular investment domain and asset class. Your responsibility is to assess the coherence, reliability, and macro-consistency of their respective recommendations in order to determine whether to adopt or reject their guidance.

\textbf{Expert Profiles:}

\begin{itemize}
    \item  \textbf{Warren Buffett}: A value investor with a long-term horizon, primarily focused on large-cap equities with durable competitive advantages. Assigned Asset: \textbf{Apple Inc. (AAPL)}
\item \textbf{George Soros}: A macroeconomic and reflexivity-oriented investor. Specializes in trading global dislocations and regime shifts. Assigned Asset: \textbf{Gold Spot Market(XAUUSD)}
\item \textbf{Peter Lynch}: A growth-oriented investor known for bottom-up stock selection and identifying fast-growing companies. Assigned Asset: \textbf{Tesla Inc. (TSLA)}
\item \textbf{Benjamin Graham}: A fundamental value investor, originator of intrinsic valuation and margin-of-safety principles. Assigned Asset: \textbf{Exxon Mobil Corporation (XOM)}

\end{itemize}

\textbf{Inputs Provided:}
\begin{itemize}
    \item \textbf{Macroeconomic Context} – A structured summary of the prevailing macro environment, including economic indicators, policy trends, market sentiment, and geopolitical signals:
\texttt{\{macro\_events\}}
    \item \textbf{Expert Recommendations} – Buy/Sell/Hold recommendations made by the four experts for their respective assets:
\texttt{\{decisions\}}

\end{itemize}
\textbf{Task Instructions:}
\begin{itemize}
    \item For each expert, analyze whether their recommendation is \textbf{contextually valid}, based on:
    \begin{itemize}
        \item  Their specific investment philosophy and historical behavioral traits
        \item  The macroeconomic landscape and potential asset sensitivity
        \item  Cross-asset interdependencies and systemic market risks
    \end{itemize}
    \item Use the following decision rules:
    \begin{itemize}
        \item  Respond **Yes** if the expert’s recommendation is well-aligned with both the macroeconomic context and their own philosophical model of decision-making.
        \item  Respond **No** if there is misalignment, excessive risk exposure, or evident inconsistency with current conditions.
    \end{itemize}
    \item Your final output should reflect \textbf{discrete endorsement decisions}, not nuanced analysis. These decisions are intended to drive automated position sizing and portfolio response.

\end{itemize}
\textbf{Output Format (strictly enforced):}

Return a single-line output with no additional explanation, using the format below:

[Buffett:<Yes or No>, Soros:<Yes or No>, Lynch:<Yes or No>, Graham:<Yes or No>]
\end{tcolorbox}

\caption{Prompt Template for Expert Alignment Agent Financial Analysis and Decision}
\end{figure*}

\end{document}